\documentclass[journal]{IEEEtran} 

\usepackage{amsmath,amsfonts}
\usepackage{algorithmic}
\usepackage{algorithm}
\usepackage{array}
\usepackage{textcomp}
\usepackage{stfloats}
\usepackage{url}
\usepackage{verbatim}
\usepackage{graphicx}
\usepackage{cite}

\usepackage{mathptmx} 
\usepackage{amssymb}  
\usepackage{hyperref}
\usepackage[table,dvipsnames]{xcolor}
\usepackage[font=small]{caption}
\usepackage{booktabs}
\usepackage{blindtext}
\usepackage{subcaption}
\usepackage{multirow}

\DeclareMathAlphabet{\mathcal}{OMS}{cmsy}{m}{n}

\definecolor{MyGreen}{rgb}{0,0.65,0}
\definecolor{LightCyan}{rgb}{0.80,1,1}
\definecolor{LightGray}{rgb}{0.87,0.87,0.87}
\definecolor{PaleYellow}{rgb}{1,1,0.85}
\definecolor{PaleGreen}{rgb}{0.80,1,0.80}
\definecolor{LightRed}{rgb}{1,0.5,0.5}
\definecolor{MyGreen}{rgb}{0,0.65,0}
\definecolor{SteelBlue}{rgb}{0.27,0.51,0.71}
\definecolor{LightSkyBlue}{rgb}{0.69,0.77,0.87}
\definecolor{LightBlue}{rgb}{0.68,0.85,0.90}
\definecolor{LightSteelBlue}{rgb}{0.53,0.81,0.98}
\definecolor{DarkGreen}{rgb}{0,0.65,0}
\definecolor{firebrick}{rgb}{0.7, 0.13, 0.13}
\definecolor{lava}{rgb}{0.81, 0.06, 0.13}

\begin{document}

\title{
A Suspended Aerial Manipulation Avatar for Physical Interaction in Unstructured Environments
}

\author{Fanyi Kong$^{1}$, Grazia Zambella$^{1}$, Simone Monteleone$^{2}$, Giorgio Grioli$^{2}$, Manuel G. Catalano$^{2}$ and Antonio Bicchi$^{1,2}$
\thanks{$^{1}$ Centro di Ricerca E. Piaggio e Dipartimento di Ingegneria dell'Informazione, Universit\`{a} di Pisa, 56126 Pisa, Italy, {\tt\small kong.fanyi95@gmail.com}.} 
\thanks{$^{2}$ Fondazione Istituto Italiano di Tecnologia, Via Morego 30, 16163 Genova, Italy.}
%
}

\graphicspath{{./images/}}

\maketitle

\begin{abstract}
This paper presents an aerial platform capable of performing physically interactive tasks in unstructured environments with human-like dexterity under human supervision. 
This aerial platform consists of a humanoid torso attached to a hexacopter.  A two-degree-of-freedom head and two five-degree-of-freedom arms equipped with softhands provide the requisite dexterity to allow human operators to carry out various tasks.
A robust tendon-driven structure is purposefully designed for the arms, considerably reducing the impact of arm inertia on the floating base in motion. In addition, tendons provide flexibility to the joints, which enhances the robustness of the arm preventing damage in interaction with the environment.
To increase the payload of the aerial system and the battery life, we use the concept of Suspended Aerial Manipulation, i.e., the flying humanoid can be connected with a tether to a structure, e.g., a larger airborne carrier or a supporting crane.  Importantly, to maximize portability and applicability, we adopt a modular approach exploiting commercial components for the aerial base hardware and autopilot, while developing an outer stabilizing control loop to maintain the attitude, compensating for the tether force and for the humanoid head and arm motions. The humanoid can be controlled by a remote operator, thus effectively realizing a Suspended Aerial Manipulation Avatar. The proposed system is validated through experiments in indoor scenarios reproducing post-disaster tasks. 

\end{abstract}


\section{Introduction}\label{sec:Intro}

\IEEEPARstart{T}{he} increasing prominence of aerial robots in the realm of mobile robotics is propelled by recent advancements in unmanned aerial vehicle (UAV) technology. The inherent capability of flight provides advantages not present in conventional ground-based robotic platforms, facilitating rapid deployment in elevated terrains, post-catastrophe environments, and other distant and hazardous locations. 
Beyond passive sensing, researchers actively explore the potential of UAVs for environmental interaction. The amalgamation of a manipulator or end-effector with a robotic arm, coupled with the inherent aerial maneuverability of a UAV platform, leads to the conceptualization and realization of Unmanned Aerial Manipulators (UAMs). 

UAMs have undergone scrutiny for various applications, including industrial pipeline inspection \cite{8435987}, wind turbine blade cleaning \cite{aerones}, and tasks involving torsion, such as screwing bolts, replacing light bulbs, and crop harvesting\cite{7759258}. The versatility of UAMs extends to potential applications in post-disaster response scenarios.
A comprehensive assessment of the evolution and prevailing trends in UAMs is available in \cite{ollero2021past}. Furthermore, \cite{ruggiero2018aerial} proposes a review of design techniques, while the authors of \cite{meng2020survey} present a survey covering mechanics, modeling, and control architectures pertinent to UAMs.


Within the realm of UAMs, dual-arm aerial platforms constitute a substantial branch. Key advantages of these platforms include heightened dexterity and enhanced manipulation capabilities resulting from an expanded workspace. Furthermore, the dual-arm structure offers the potential to partially eliminate reaction wrenches on the platform through compensatory movements, as illustrated in \cite{suarez2018design}.
Table \ref{tab:dual_arm_manipulators} compiles recent dual-arm aerial platforms, presenting a comparative analysis of their primary technical specifications. This analysis encompasses parameters such as arm weight, payload capacity, Degrees of Freedom (DoFs) of the arms, and the incorporation of compliance features.
In \cite{suarez2018design} (row 1 of the Table \ref{tab:dual_arm_manipulators}), the authors present a hexacopter-based aerial manipulator with a lightweight arm design implemented with an aluminum frame.
In \cite{suarez2018access} (row 2), compliance is introduced into the lightweight dual-arm design by incorporating a compact spring-lever transmission mechanism.
The authors in \cite{app10248927} (row 3) investigate bimanual aerial manipulation tasks with one arm grabbing to a fixed point on a lightweight and compliant dual-arm aerial manipulation robot.
An open-source, low-cost dual-arm system for aerial manipulation is presented in \cite{perez2020hecatonquiros} (row 4).
The authors of \cite{machines10040273} (row 5) present a dual-arm UAM with an anthropomorphic design and successfully assemble two workpieces in an outdoor experiment.
The authors of \cite{8059875} (row 6) and \cite{korpela2014towards} propose an aerial platform with two 4-DoF arms and apply it to valve turning.
In \cite{prodrone} (row 7), the readers can find a commercial version of a dual-arm aerial manipulator.


To avoid collision between UAVs and the obstacles in complex environments and dynamic turbulence caused by ground effects, long reach configuration has been introduced in aerial manipulator design. 
The long reach configuration involves suspending the robotic manipulator from the UAV using a long rigid or flexible link (e.g., long wires), rather than fixing it directly to the UAV.
\cite{8404953}, \cite{8593940} and \cite{9088973} presents several aerial manipulators prototypes that attach a lightweight dual arm to the aerial platform in long reach configuration via a passive link and their industrial applications in the pipeline inspection and sensor device installation. \cite{9571044} presents a long reach manipulator suspended by two strings applied to the installation of helical bird diverters on power lines.
In these prototypes, the manipulators are passively suspended from the drones like pendulums. The swinging motion naturally introduced is seen as a disturbance that is compensated for by the drone's control.
In \cite{8743463}, the authors install non-vertical ducted fans on the suspended gripper module to suppress the string swing effect, 
In addition, they used a winch to control the distance between the gripper and the multi-rotor platform.
The authors of \cite{8793592} present a similar cable-Suspended Aerial Manipulator (SAM) actuated by winches and non-vertical propulsion units, and propose an oscillation damping control in \cite{9197055}. When analyzing the system model, the authors of these two prototypes considered the aerial manipulator independently of the vehicle, assuming it is suspended from a fixed point and suppressed its swing by controlling the propulsion device.

Teleoperation, as a prevalent method of robot control, enables human supervision and intervention in task execution through a master-slave mode. Researchers have explored its application in aerial manipulators as well.
The authors of \cite{coelho2021whole} introduce a passivity-based control framework for the teleoperation of a kinematically-redundant aerial platform. 
The aerial base is equipped with a monocular camera, and its perspective is adjusted using the platform redundancy. Task execution involves switching among control options facilitated by a 2-DoF joystick.
In \cite{9197394}, the authors introduce a telepresence system employing visual-inertial feedback to construct a real-time virtual representation of the workspace for the operator within virtual reality (VR). 
Similarly, the authors of \cite{8981574} and \cite{9476884} present comparable teleoperation system designs, featuring the presentation of a digital twin of the aerial robot and the remote environment in virtual reality to the user through a head-mounted display. Notably, in \cite{8981574}, manipulator movement is controlled by trackers worn by the operator, while in \cite{9476884}, a set of joysticks serves as the input device.

\begin{table*}[t]
    \caption{Examples and primary characteristics of dual-arm aerial platforms in the literature.}
    \label{tab:dual_arm_manipulators}
    \centering
    \begin{center}
    {\setlength{\extrarowheight}{1pt}
        \begin{tabular}{|m{1.1cm}| m{1.3cm}| m{1.4cm}| m{1.0cm}| m{0.6cm}| m{1.2cm}| m{1.4cm}| m{1cm}| m{1.2cm}| m{1.2cm}| m{1.6cm}|}
        \hline     
        \rowcolor{LightSkyBlue}
        {Reference} & {Aerial Base} & {Arm Weight} & {Arm Payload (each)} & {Arm DoF (each)} & {Compliant} & {End Effector} & {Head} & {Open-Source} & {Simulator} & {Teleoperation} \\
        \hline
        \cite{suarez2018design} & Hexacopter & 1.8 kg total& 0.75 kg & 5 & No & Gripper & Fixed Camera & No & MATLAB & 6-DoF mouse \\
        \rowcolor{LightGray}
        \cite{suarez2018access} & Hexacopter & 1.3 kg total& 0.2 kg & 4 & Yes & Gripper & Fixed Camera & No & No & 6-DoF Mouse \\

        \cite{app10248927} & Hexacopter & 1.0 kg total & 0.3 kg & 3 & Yes & Gripper & No & No & No & No \\

        \rowcolor{LightGray}
        \cite{perez2020hecatonquiros} & Hexacopter & 0.5 kg each & NA & 4 & No & Multiple Gripper & Fixed Camera & Yes & OpenRave & Gamepad, Joystick, GUI \\
        \cite{machines10040273} & Hexacopter & 1.9 kg total & NA & 4 & No & Gripper & Fixed Camera & No & MATLAB & No \\

        \rowcolor{LightGray}
        \cite{8059875} & Quadrotor & NA & NA & 4/2 & No & Gripper & No & No & No & No \\

        \cite{prodrone} & Hexacopter & NA & NA & NA & No & Gripper & NA & Commercial & No & NA \\
        \rowcolor{LightGray}
        \textbf{proposed} & Hexacopter & 2.85 kg each & 2 kg & 5 & Yes & SoftHand\cite{catalano2014adaptive} & Movable & Yes & Gazebo & VR \\
       \hline
        \end{tabular}}
    \end{center}
\end{table*}

\begin{figure}[t]
   \centering
   \includegraphics[width = 0.9\linewidth]{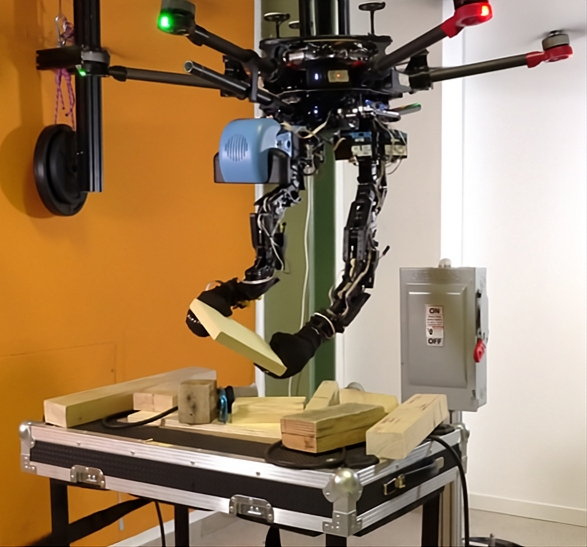}
   \caption{Prototype of the aerial platform used for aerial teleoperation. It is composed of a DJI hexacopter base, two 5-DoFs arms based on novel soft articulated joints, two softhads (derivation of the PISA/IIT SoftHand \cite{catalano2014adaptive}), and a 2-DoFs head.}
   \label{fig:Aerial_Platform}
\end{figure}

In our previous work \cite{9981910}, on which this manuscript is partially based, we have introduced a novel design of a hexacopter equipped with an anthropomorphic torso comprising two soft-articulated 5-DoF arms, two softhands, and a 2-DoF head. 
A notable innovation in the arm design has been introduced, involving the integration of tendons and elastic bands. It permits the positioning of the actuation unit away from the rotation shafts and increases the robustness of the system against accidental shocks.
The softhands \cite{catalano2014adaptive} have been adapted to match the aerial platform design to enhance the grasping and manipulation performance, owing to their capacity to conform the grasp to the shape of objects.
A stereo camera mounted on a compliant two-degree-of-freedom head has been used for perception.

Here, we extend \cite{9981910} by integrating the aerial platform design with a variable-length suspension mechanism for weight compensation (see Fig.~\ref{fig:Aerial_Platform}), effectively making it a Suspended Aerial Manipulation Avatar (SAM-A). 
This design, inspired by \cite{8793592}, considerably extends the system's flight duration and endows it with the versatility to attach to various carriers such as cranes or manned aerial vehicles, facilitating access to areas where UAV operations may be restricted. 
However, our approach overcomes some of the limitations of \cite{8793592} by introducing non-fixed compensation to preserve freedom of motion for the aerial platform.
Additionally, we present an aerial base stabilizing controller that takes into account the motion of the humanoid torso and the influence of the tethering system. This controller is integrated within a teleoperation framework, enabling human operators to control the platform in an immersive and intuitive manner.
Moreover, our system adopts a modular approach, utilizing a commercial UAV as the aerial base and offering open-source design, providing an easily replicable solution and contributing to the portability and applicability of our aerial platform modeling and control design to similar configurations.
The efficacy of the proposed controller is validated through experiments conducted within an indoor test rig. A series of representative experiments simulating scenarios relevant to post-disaster reconstruction efforts are conducted to demonstrate the functionality and effectiveness of our design.



We organized the paper as follows. Section \ref{sec:Concept} explains the motivation and provides an overview of our system.
Section \ref{sec:Design} covers the mechatronic design of the system.
Section \ref{sec:MODEL} describes the system model.
Section \ref{sec:Control} presents the control architecture. 
Finally, Section \ref{sec:Results} displays the experimental setup and results and Section \ref{sec:Conclusion} concludes the paper.



\section{Concept} \label{sec:Concept}

After a disaster such as an earthquake, a flood, a fire or a landslide, there is often the need for robots to access high-risk environments as observers and operators. 
Ground robots can be hampered by debris and objects on the ground, and safely surpassing them is not always trivial.
On the contrary, a robot capable of approaching mission areas from the air is intrinsically immune to such obstacles and, therefore, greatly enhances the possibility of inspecting and intervening in adverse surroundings for humans.
However, limited flight time and payload, as well as unknown challenges and potential threats in the unstructured working space, are some of the notable constraints limiting the application of aerial robots in practical scenarios.

In addition to relying on advancements in battery technology for extending flight time, a common strategy involves reducing the weight of the aerial manipulator during the mechanical design phase. Researchers often achieve this by selecting low-density, high-strength materials and navigating a compromise between degrees of freedom, functionality, and overall mass.
A notable departure from conventional non-grounded robots is the concept of a cable-suspended aerial manipulator, introduced in \cite{8793592}. In this paradigm, an aerial manipulator is designed to be suspended from an aerial carrier, such as a manned aerial vehicle or a crane. The carrier bears a significant portion of the manipulator weight, leading to a substantial increase in payload capacity and aligning with the overarching objective of prolonging flight time.

UAVs are inherently unstable systems, relying on continuous and substantial rotor efforts to maintain stability.
Compared with ground robots, the disturbances to aerial manipulators brought by operations may have a more devastating impact, making it particularly challenging to conduct interactive tasks in unknown environments where there are more potential disturbances.
Therefore, when designing aerial manipulators, particular attention must be dedicated to minimizing those disturbances caused by arm movement and interaction.
One of the lessons in the arm design with application to aerial manipulation is to constantly maintain their Center of Mass (CoM) close to the CoM of the UAV. 
Consequently, it is imperative that the base of the arm carries the majority of its weight.
Additionally, compliance can be integrated into the design, which can prevent the arm from breaking during accidental collisions between the robot and the environment and mitigate the impact on the stability of the UAV caused by sudden and impulsive loads encountered during manipulation. 
Furthermore, in order to tactfully respond to various situations that may arise in unknown environments, a practical strategy is to maintain human supervision of the aerial manipulator rather than relying entirely on its autonomy.

\subsection*{Proposed Approach}

\begin{figure}[t]
    \centering
        \includegraphics[width = 1.0 \linewidth]{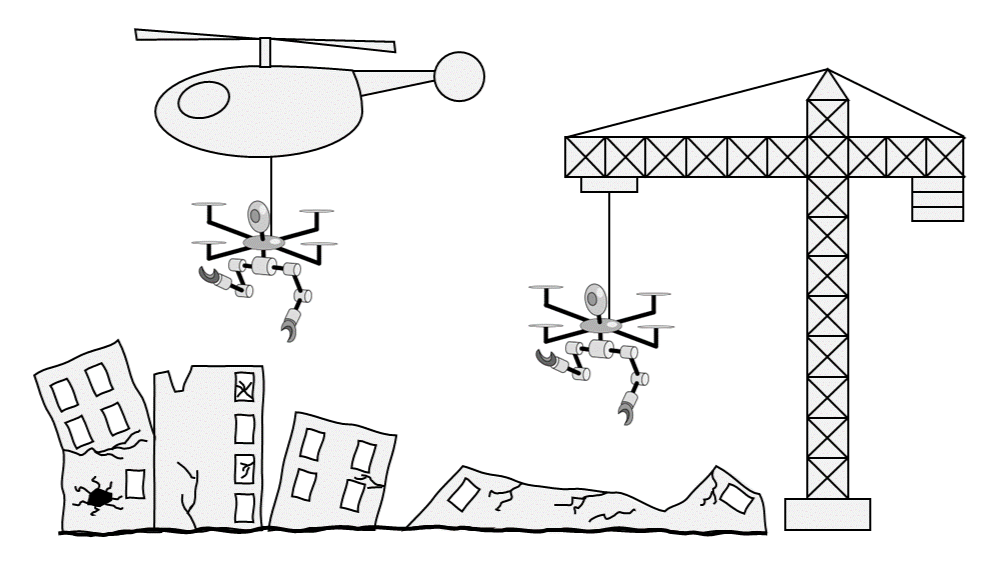}
        \caption{Concept of SAM-A.
        The Suspended Aerial Manipulation Avatar is designed for operation in highly unstructured environments, such as after a disaster. An external support device such as a helicopter (left) or a crane (right) improves payload bearing and autonomy, while an independent propulsion system stabilizes the bi-manual manipulating platform.        
        }\label{fig:Concept SAMA}  
\end{figure}

Taking inspiration from SAM \cite{8793592} and the aerial manipulators described in Table \ref{tab:dual_arm_manipulators}, we aim to design a Suspended Aerial Manipulation Avatar for inspection and intervention in unstructured environments in which human intelligence is relocated directly in the aerial platform working space (see Fig.~\ref{fig:Concept SAMA}).

In contrast to the approach proposed in \cite{8793592}, which relies entirely on the tether for lifting the entire weight of the robot but confines its motion to a spherical surface, we address the challenges of flight duration and load capacity through the implementation of a variable-length suspension system for gravity compensation. This system partially offsets the overall weight, enabling the aerial base to utilize its thrusters for various maneuvering operations, including ascending, descending, and translating. This design preserves the inherent characteristics of the aircraft and significantly upholds the independence of the aerial base from the cable suspension system. Consequently, it allows for movement within a larger workspace and facilitates more agile operations by coordinating the aerial base motion with the arm motion, rather than relying solely on the degrees of freedom of the arms.

The arms and head are affixed to the underside of the UAV, configured to mimic the upper body of a human but with a downward orientation.
The dual arms are symmetrical to reduce the burden on the rotors. A tendon-elastic-band-pulley structure is applied to the arm design to meet the requirements given by stability and robustness.
This design choice allows the actuation units, which contribute significantly to the overall weight, to be situated in close proximity to the body on the aerial platform. Moreover, the joints comprised of disarticulated components are intentionally designed to be less rigid owing to the incorporation of tendons and elastic bands.
A similar approach to compliance is adopted in the head joints to mitigate the potential accidental impacts. This design strategy enhances the overall adaptability and safety of the aerial manipulator system.
Finally, two SoftHands (SHs) are employed as end-effectors. The SHs have a unique design that consents to many dexterous grasps with a sole actuation unit. When grabbing objects with conventional shapes, standard grippers would be more reliable. The SHs, on the other hand, proved to be more adaptable to performing the various grasps \cite{catalano2014adaptive} resulting from employing the aerial manipulator in unstructured environments in which the shape of objects is not known in prior.
Moreover, the mechanics of the hands are intrinsically robust against accidental collisions, making the use in teleoperation safe and effective.

In contrast to many of the aerial manipulators detailed in Table \ref{tab:dual_arm_manipulators} and \cite{coelho2021whole} where the camera is fixed to the aerial base, our design incorporates a humanoid head with two degrees of freedom rotation. This configuration decouples the perspective of the camera from the movement of the aerial base, enhancing the observation of the surrounding situation.
The platform is equipped with a binocular camera, supplemented by a VR headset on the ground station to recreate the 3D perspective of the robot for the human operator.
This first-person, real perspective offers more intuitive information about the workspace, including depth information, compared to the third-person perspective of digital twins presented in \cite{9197394}, \cite{8981574}, and \cite{9476884}.
Drawing inspiration from \cite{lentini2019alter} and \cite{park2018interactive}, we employ two handheld joysticks to track the movement of the operator's hands, controlling the robotic arms to follow and replicate human movements.
The first-person visual perception and control make teleoperation intuitive and straightforward, providing human operators with an immersive experience akin to being in the environment where the robot operates, allowing them to handle various tasks and respond to emergencies effectively.
The combination of this teleoperation experience, along with the humanoid robotic torso equipped with dexterous hands, effectively positions this aerial robotic platform as an "Avatar."

The aerial base employed is a conventional commercial hexacopter equipped with parallel thrusters, with the proposed control system designed to run on top of it. While its under-actuated nature may introduce complexity in decoupling and stabilization compared to fully-actuated drones, it offers the distinct advantages of simplicity and reliability. Moreover, the technology associated with such hexacopters is thoroughly researched and extensively commercialized, enhancing the feasibility of our proposed solution for widespread application across various scenarios.

\section{Platform Design} \label{sec:Design}





\begin{figure*}[ht] 
\centering
\subfloat[]{\label{fig:joint_}
\includegraphics[width=0.72\columnwidth]{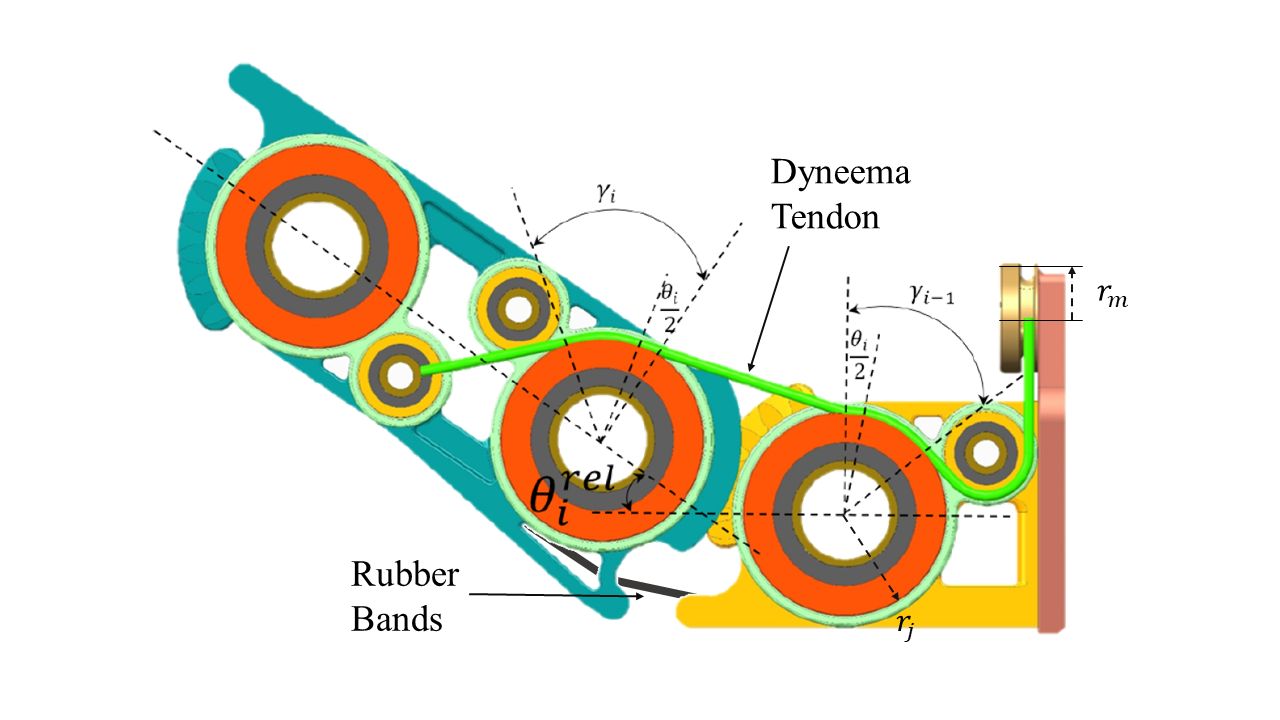}}
\hfil
\subfloat[]{\label{fig:joint_routing}
\includegraphics[width=0.32\columnwidth]{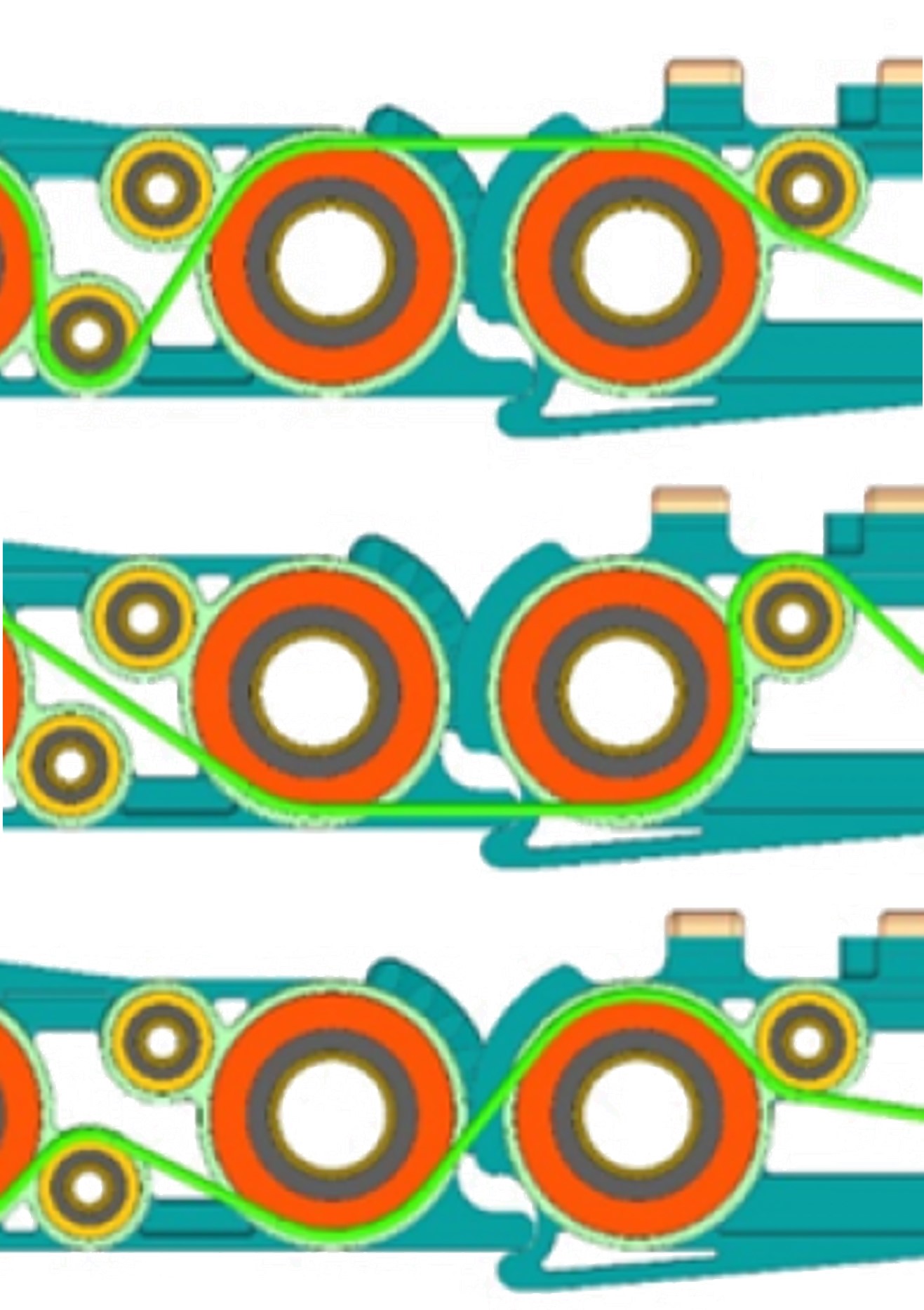}}
\hfil
\subfloat[]{\label{fig:tendon_config}
\includegraphics[width=0.92\columnwidth]{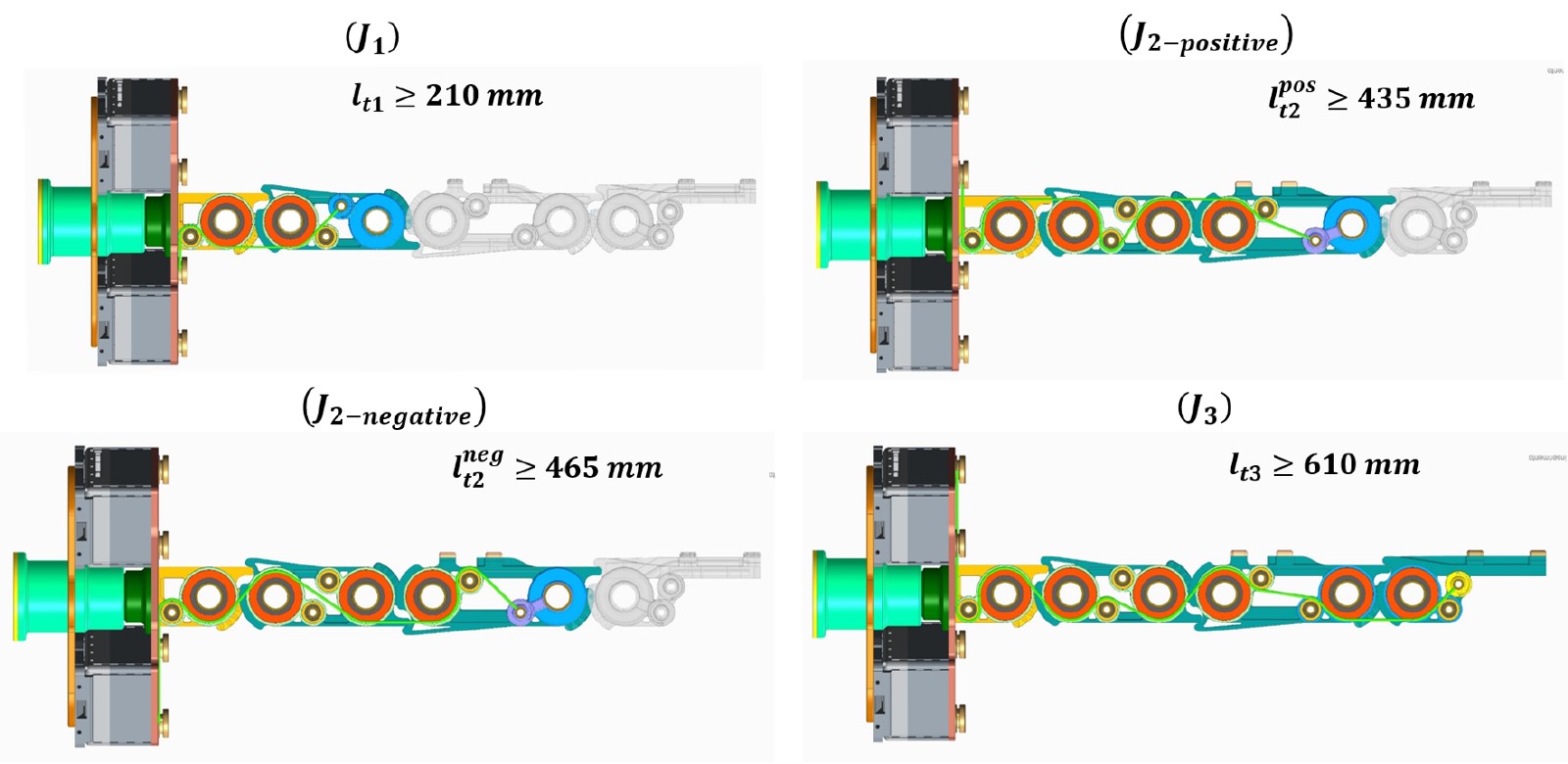}}
\caption{a) Configuration of the joints. The figure shows the kinematic relations between the motor rotation and the joint rotation and presents the two axes around which the joint rotates. The elastic bands keep the joint in position while The tendon drives it. b) Routing of the tendons inside a joint. From top to bottom, positive configuration closes the joint, negative configuration opens it, and neutral configuration exerts no action on it. c) Routing of the four tendons to achieve the differential actuation of the joints.}
\label{fig:Arm_joint_tendons}
\end{figure*}


\subsection{Mechanical Design} \label{sec:mechanical_design}
As discussed above, the aerial avatar is designed for accessing unstructured environments and manipulating objects. The main mechanical components of the systems are the hexacopter aerial base, which constitutes the body of the platform, the arms, the hands, the head, and the compensation system. The open-source materials are available online\footnote{https://www.naturalmachinemotioninitiative.com/aerial-alter-ego}.

\subsubsection{Aerial Base}
While the tether can compensate for a significant portion of the weight of the aerial manipulator, enhanced drive capabilities are still necessary for the aerial base. This requirement translates to faster response times and improved dynamic performance. To enhance portability and broaden the applicability of the system, the aerial base is constituted by a DJI Matrice 600 Pro\footnote{https://www.dji.com/it/matrice600-pro}.
The hexacopter weighs 9.5 kg and lifts a maximum recommended takeoff load of 15.5 kg. Therefore, the net recommended payload is 6 kg. There is also the possibility of temporarily over-boosting the propellers, increasing the maximum load to 10 kg.
The dimensional encumbrance in operative working conditions is 1668 mm $\times$ 1518 mm $\times$ 727 mm with propellers, frame arms, and GPS antennas unfolded.


\subsubsection{Dual Arms}

The arm design principle is governed by three primary considerations: dexterity, total weight, and weight distribution. While increased degrees of freedom contribute to greater dexterity, they concurrently add weight to the arms, introducing disadvantages for aerial base control. Thus, striking a balance between dexterity and total weight is generally necessary. Furthermore, the distribution of weight must be taken into account to prevent substantial changes in the CoM and minimize interference moments on the aerial base resulting from arm movements.

To achieve these goals, the arm design is inspired by the tendon-driven manipulators that exist in the literature \cite{wang2021survey}, particularly the phalanges of the IIT/Pisa SoftHand \cite{catalano2014adaptive}. It introduces tendon-driven actuation with a tensegrity structure implemented with elastic cables into the elbow design, and presents two main advantages.
First, unlike conventional robotic arms with direct transmission such as the ones in \cite{suarez2018design} that place the motors at the joints, the tendon design allows most of the motors, which account for considerable weight, to be placed at the base of the arm. This design optimizes weight distribution for aerial manipulator applications, bringing the CoM of the arm closer to the aerial base and reducing the weight of the arm links, thereby minimizing its backlash on the body.
The latter is that the elastic bands present soft ligaments to the joints, which improves the robustness of the arm and protects it in the case of accidental collisions \cite{4660316}.
Based on this design, the elbow is actuated by tendons and has three rotational DoFs to reduce CoM shifting during arm grabbing and lifting, thereby further improving stability.
The shoulder and wrist joints are driven independently by two motors located at the base of the arm and at the wrist.
In total, the arm has five DoFs, including one for the shoulder, three for the elbow, and one for the wrist. The kinematic definition of the arm is reported in Table~\ref{tab:DHChain}.

\begin{table}[htbp]
    \caption{Denavit-Hartenberg chain of each arm.}
    \label{tab:DHChain}
    \setlength{\tabcolsep}{10pt}
    \centering
    \begin{tabular}{|c|c|c|c|c|}
    \hline  
    \rowcolor{LightSkyBlue}
         Link & $a$ & $\alpha$ & $d$ & $\theta$  \\
         \hline
         1 & $0$ & $-\pi/2$ & $0$ & $-\pi/2 + \theta_1$ \\
         \rowcolor{LightGray}
         2 & $0.110$ & $0$ & $0$ & $-\pi/2 +\theta_2$ \\
         3 & $0.140$ & $0$ & $0$ & $\theta_3$ \\
         \rowcolor{LightGray}
         4 & $0$ & $\pi/2$ & $0$ & $\pi/2 + \theta_4$ \\
         5 & $0$ & $0$ & $0.220$ & $\theta_5$ \\
         \hline  
    \end{tabular}
    \vspace{-0.0cm}
\end{table}

As shown in Fig. \ref{fig:joint_}, the joint is created with two pulleys and toothed profiles. 
A tendon made of Dyneema fiber is wrapped around these two pulleys and connected to a motor at the bottom of the arm. It deforms less than one percent at nominal load and is strong enough to drive the arm linkage to follow the toothed profile of the pulleys as the motor rotates, resulting in a moving center of rotation.
A set of rubber elastic bands contributes to the structural integrity, ensuring that the toothed profiles stay in position during normal rotation and under accidental impacts.
The transmission ratio from the motor to the shaft joint is
\begin{equation}
   r = \tfrac{r_m}{r_j} \, ,
\end{equation} 
where $r_j$ is the radius of the joint pulley, $r_m$ is the radius of the pulley connected to the motor.
It is worth noticing that a motion of the elbow joint $\theta_i$ is achieved by two separate rotations.
Following a motor movement that produces a motion of $\theta_i$, the former is generated by rotations of $\theta_i/2$ along two different axes of the pulleys constituting a joint. 

The design incorporates a system where a single tendon can only exert pull force, necessitating the use of at least two tendons to enable movement in two directions for one joint. Three configurations of tendon application are employed when wrapping around a joint, as illustrated in Figure \ref{fig:joint_routing}. In the positive configuration, the tendon closes the joint; in the negative configuration, it opens it, while in the neutral configuration, it merely passes through the joint without generating any torque.

Utilizing these configurations, four tendons drive three joints of the elbow through a differential arrangement, as depicted in Figure \ref{fig:tendon_config}. The first tendon is positively wrapped around the first elbow joint $J_1$. Conversely, the second tendon passes around $J_1$ in a negative arrangement and, simultaneously, positively around $J_2$. The third cable is neutral on $J_1$ and negative on $J_2$. Finally, the fourth tendon is neutral around $J_1$ and $J_2$, and positive around $J_3$. The closing action on $J_3$ is facilitated by elastic rubber cables and is thus passive. The elastic behavior on the joint is experimentally evaluated through traction tests, approximating a linear dependence between force and elongation within the working range.
When appropriately controlled, this tendon displacement configuration enables independent movement of all the joints.

The kinematic projection between the arm joints $q_{a}$ ($q_{ra}$ for the right arm and $q_{la}$ for the left) and the drive motors $q_m \in \mathbb{R}^{6}$ is
\begin{equation}
    q_{a} =  F \cdot q_m \label{eq:Conf_Matrix} \, ,
\end{equation} 
where
\begin{equation}
    F = \begin{bmatrix}
    1& 0 & 0 & 0 & 0 & 0 \\
    0 & r & -r & 0 & 0 & 0 \\
    0 & 0 & r & -r & 0 & 0\\
    0 & 0 & 0 & 0 & r & 0  \\
    0 & 0 & 0 & 0 & 0 & 1 \\
    \end{bmatrix} \, .
\end{equation} 
 
\begin{figure}[h]
  \centering
    \includegraphics[width = 0.9 \linewidth]{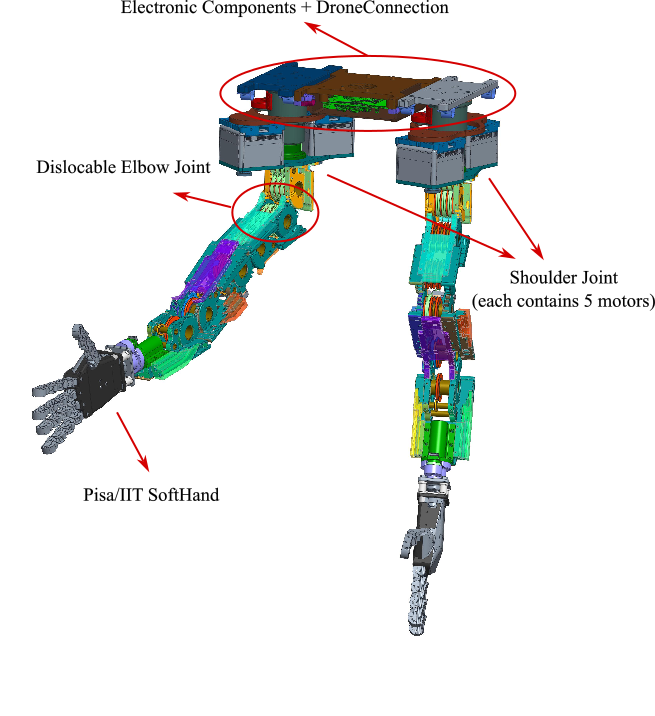}
  \vspace{-0.7cm}
  \caption{Description of the torso of the platform. The picture points out the position of the electronics, the localization of the actuation units of both arms, the disarticulated joints, and the SohtHands employed as end-effectors.}
  \label{fig:arm_design}
  \vspace{0cm}
\end{figure}

The arm links are made by 3D printing with a lightweight hollow frame design. Fig. \ref{fig:arm_design} shows the dual arm assembly. The whole arm weighs approximately 2.85 kg, divided by around 2.0 kg on the shoulder and 0.85 kg on the rest of the arm (including a 0.35 kg hand). Ideally, without considering the aerial base, the end of the arm can output 6 Nm of maximum torque and 30 N of maximum force.

%

\subsubsection{Hands}
For each arm, a lightweight version of PISA/IIT Softhand \cite{catalano2014adaptive} is equipped for manipulation.
It is an optimal solution that compromises weight, dexterity, and compliance for aerial manipulation applications.
The SoftHand is a 19 DoF system weighing around 0.35 kg, which is actuated by a single motor and a tendon that wraps around all the fingers, allowing maximum holding torque of 2 Nm and maximum holding force of about 20 N.
Their employment enhances the aerial avatar capabilities to handle objects of different shapes and dimensions, thanks to their intrinsic adaptability and softness.
For more information, the authors suggest referring to  \cite{catalano2014adaptive}.

\subsubsection{Head Design}
\begin{figure}[h]
  \centering
  \includegraphics[width = 0.5 \linewidth]{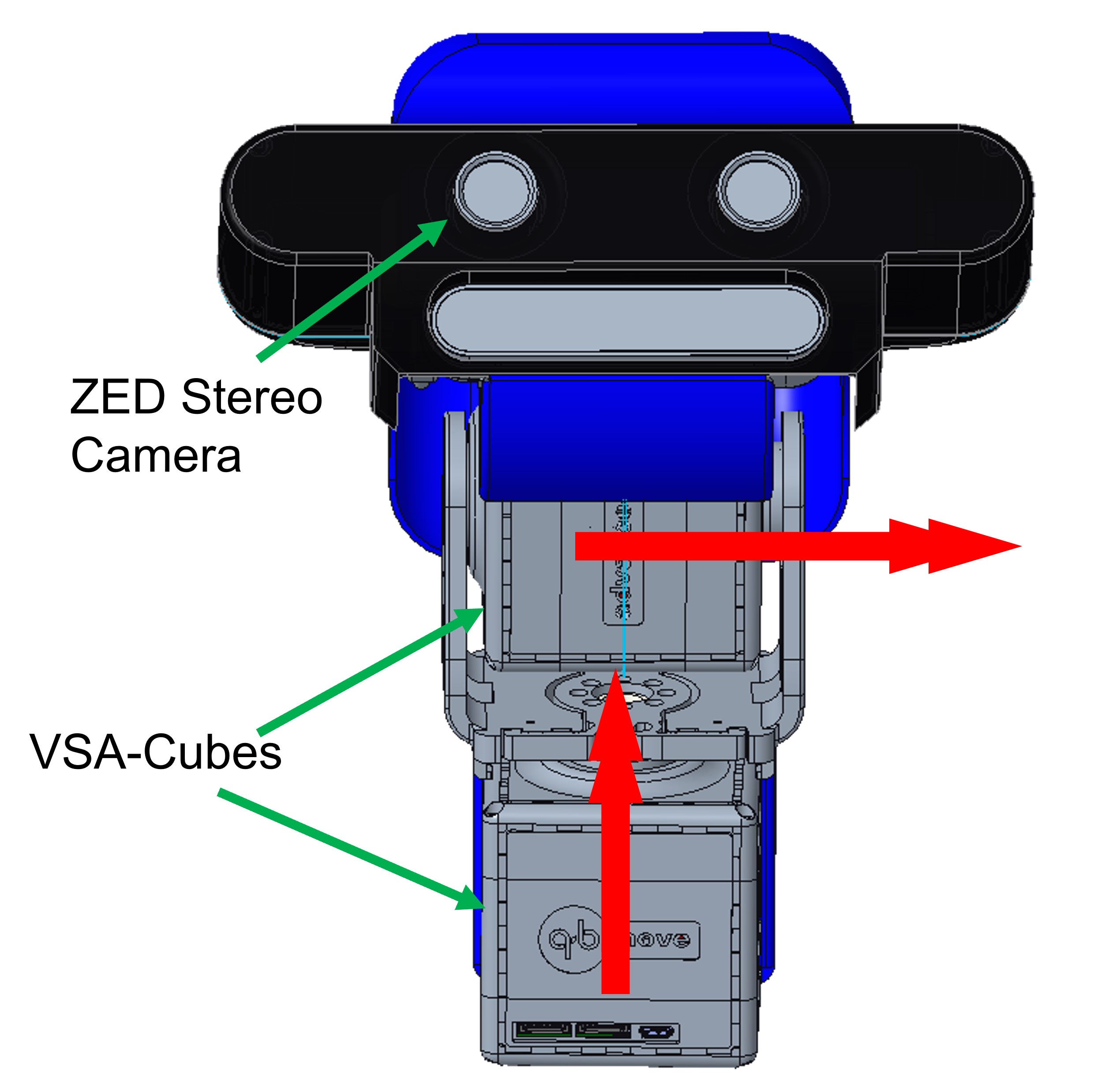}
  \caption{2-Dofs head, showing the direction of rotation and the position of the binocular camera.}
  \label{fig:head_design}
\end{figure}
The head is located at the bottom of the aerial base and in the middle of the two arms, installed on the landing gear of the UAV through a four-bar linkage.
The head design is similar to the one already employed for Alter-Ego \cite{lentini2019alter}. 
It consists of a 2 DoFs neck and a Zed Mini Stereo Camera to "see" the workspace while teleoperating (see Fig. \ref{fig:head_design}).
The neck joints employ two VSA-cubes \cite{catalano2011vsa} as actuation units, which introduce compliance to cope with possible impacts.
The neck allows rotating the head around the direction of pitch and yaw, as its kinematics model defined in Table~\ref{tab:DHChain_Head}.
The choice is related to the sensations that an operator experiences while commanding such robotic platforms. Experimentally, we found out that yaw and pitch were fundamental to avoid the sense of nausea and disorientation during teleoperations. 
Conversely, the absence of a rolling movement is not experienced as stressed by operators; therefore, we eliminated it to reduce the head weight to approximately 1.5 kg.

\begin{table}[htbp]
    \caption{Denavit-Hartenberg chain of the head.}
    \label{tab:DHChain_Head}
    \setlength{\tabcolsep}{10pt}
    \centering
    \begin{tabular}{|c|c|c|c|c|}
    \hline  
    \rowcolor{LightSkyBlue}
         Link & $a$ & $\alpha$ & $d$ & $\theta$  \\
         \hline

         1 & $0$ & $\pi/2$ & $0.056$ & $\pi/2 + \theta_1$ \\
         \rowcolor{LightGray}
         2 & $0.094$ & $0$ & $0$ & $0.52+\theta_2$ \\
         \hline  
    \end{tabular}
    \vspace{-0.0cm}
\end{table}

\subsubsection{Compensation system}
The compensation system adopts a variable-length design, allowing the rope length to change while exerting tension on the aerial platform. It consists of a force generator and a force adjuster.
The compensation force is generated by the force generator and then adjusted by the force adjuster before being applied. 

\begin{figure}[h] 
\centering
\subfloat[]{\label{fig:compensation_1}
\includegraphics[width=0.6\columnwidth]{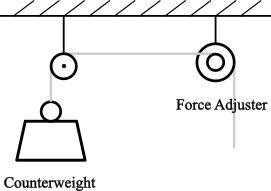}}
\hfil
\subfloat[]{\label{fig:compensation_2}
\includegraphics[width=0.6\columnwidth]{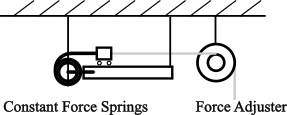}}
\hfil
\subfloat[]{\label{fig:variator}
\includegraphics[width=0.7\columnwidth]{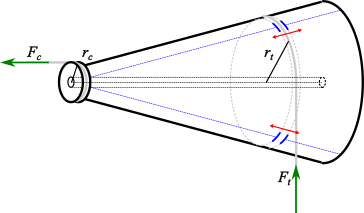}}
\caption{a) The counterweight solution utilizes a counterweight and a pulley, with a string attached to the weight and connected to the force adjustor. b) In the constant force spring solution, two constant force springs are fixed at one end of the sliding guide rail and connected to a sliding cart that can move along the guide rail. A string is tied to the other end of the cart to output the force. c) Force adjuster presents a conical design. At the top of the cone is a pulley connected to the force generator. Three worm gears are on the cone parallel to its surface and symmetrically distributed with its central axis, mounted with three pulley seats that can traverse and create a pulley with an adjustable radius. }
\label{}
\end{figure}

Considering the impact on system dynamics, two alternative configurations are designed as force generators: counterweight or constant force springs, as demonstrated in Fig. \ref{fig:compensation_1} and Fig. \ref{fig:compensation_2}.
The counterweight design involves a counterweight and a pulley. The compensation force is generated by the gravity of the counterweight and transmitted through the string and pulley to the force adjuster.
The force generated by the counterweight can be modeled as
\begin{equation}
     F_c =  m_cg + m_c a_c
 \end{equation}
 where $m_c$ is the mass of the counterweight and $a_c$ is its acceleration.
The constant force spring design comprises two constant springs, a prismatic sliding guide, and a cart. Two constant force springs are fixed on the guide and connected in parallel to one end of the cart that can move freely along the guide to form a combined force to pull the cart. The other end of the cart is connected to the force adjuster via a string.
The force generated by the constant force spring can be expressed as
\begin{equation}
     F_c =  2 F_s
\end{equation}
 where $F_s$ is the constant force of each spring.

The counterweight design introduces additional inertia into the system, potentially adversely affecting dynamics during fast motion. However, it offers simplicity and reliability, with the weight of the counterweight easily adjustable.
On the other hand, the constant force spring solution avoids introducing additional weight (assuming the mass of the cart is negligible) and provides a constant compensating force. However, this design is associated with drawbacks, including a more complex installation and fixation design, friction introduced by the sliding guide for spring installation, and the limited selection of constant force springs, which complicates the adjustment of the output force.

The force adjuster, as depicted in Fig. \ref{fig:variator}, is primarily devised for situations where the applied force cannot be arbitrarily changed, such as the weight of the counterweight or the tension of the spring. It constitutes a fixed combination of a standard pulley and a conical pulley. The conical pulley can continuously vary the wrapping radius by sliding the pulley sockets through a set of worm gears. This mechanism allows for changing the force arm and, consequently, adjusting the final output compensation force.
The force finally exerted on the platform is  
\begin{equation}
     F_t = \frac{F_cr_c}{r_t}
\end{equation}

\subsection{Electronics, Sensors, and Communication}\label{sec:electronics}
The aerial platform is equipped with two onboard computational units. The fore is an \textit{Intel NUC7 i7 BNH} that uses Linux and ROS (Robotic Operating System). It is addressed to command the system and interface with the ground teleoperation computers. 
The latter is a \textit{DJI A3 Pro Flight Controller} that manages the signals coming from the first computer and produces the related commands on the rotors.
The ground control station (GCS) comprises a Dell Alienware laptop and a set of \textit{Oculus Rift S} VR user interfaces.

On the platform, each arm consists of three custom electronic cards that command the six motors, including one DCX22S-12V motor with GPX22 83:1 gearbox for the shoulder actuation,
four DCX22s-24V motors with customized 204.8:1 gearbox for the elbow actuation and one DCX16S-12V motor with GPX10 111:1 gearbox for the wrist actuation. 
Another electronic card is used for the VSA-Cubes of the head. 
Low-level motor position control has been implemented and integrated into the electronics, and a more detailed description can be found in \cite{catalano2011vsa}.
The hands rely on their integrated electronics, with a more detailed description in \cite{catalano2014adaptive}. 

Six 24 V batteries (DJI TB47S) power the DJI aircraft in $3P2S$ configuration and provide power outputs at 48V, 24V, and 18V. 
The 48V source is responsible for powering the propellers, while the 24V is dedicated to supplying power to the arms and the head in parallel through three independent chains. Additionally, the 18V source is utilized to power the onboard computer.

The aerial platform possesses three \textit{D-RTK GNSS}, which are high-precision navigation and positioning systems that provide an accurate, centimeter-level 
3D positioning in outdoor environments. It is also capable of withstanding magnetic interference. 
Moreover, the Zed Mini Stereo Camera (Fig. \ref{fig:head_design}) embedded in the head provides the operators with visual feedback on the workspace.


The data channels of the head and arms electronics are coupled in a series and connected to the first computer via a serial port.
The stereo camera is connected to the onboard computer through an independent USB 3.0 port.
Between the GCS and aerial platform, bilateral communication is ensured by a dedicated system. The communication channel consists of a 5GHz wireless connection, which allows the exchange of the control and vision data. The wireless connection uses a WiFi bridge. However, a 5G internet key can achieve the same result, eliminating the necessity of flying near the WiFi router.
The framework is similar to the one employed in a previous work  \cite{lentini2019alter}, and we refer to it for additional details.





\section{System Model} \label{sec:MODEL}



\begin{figure*}[h] 
\centering
\subfloat[]{\label{fig_model3d}
\includegraphics[width=0.3\textwidth]{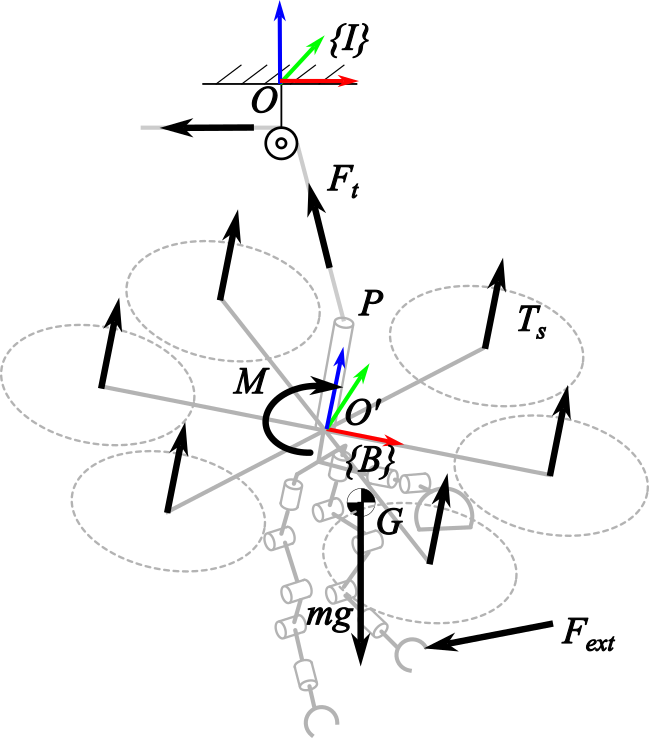}}
\hfil
\subfloat[]{\label{fig:cad_assembly}
\includegraphics[width=0.45\textwidth, trim= 2pt 2pt 0 2pt, clip]{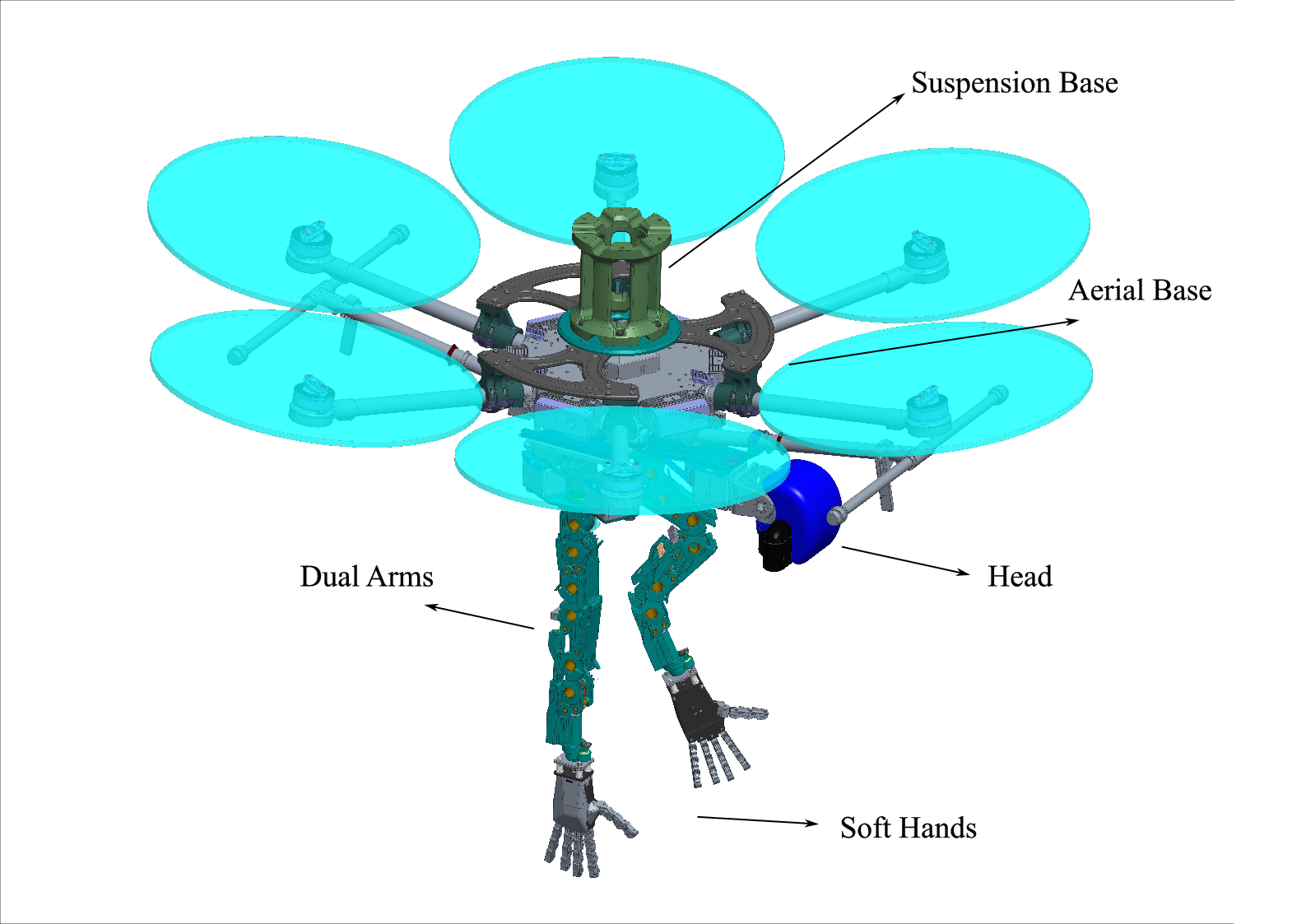}}
\caption{a) System model. b) Visualization of the humanoid design of the aerial manipulator.}
\label{}
\end{figure*}


\subsection{Kinematics}

To describe the suspended aerial manipulator model from a kinematic and dynamic point of view, we consider the airborne carrier to be hovering and therefore the suspension point is considered stationary.

Define $\{I\}$ as the inertial frame attached to it, and $\{B\}$ as the frame attached to the body with the origin located at the geometric center $O'$ of the vehicle (shown in fig.~\ref{fig_model3d}). The first and second axes of the body-fixed frame lie in a common plane defined by the six rotors, pointing forward and leftward, and the third axis is orthogonal to the plane, pointing upward of the platform.

The kinematic model of the system can be considered as a floating base mobile robot described by 
\begin{equation}
    q = \begin{bmatrix}
        q_{fb} \\ q_{lb}
    \end{bmatrix} \, ,
\end{equation}
where $q_{fb} \in \mathbb{R}^6$ is the conﬁguration of the floating base and $q_{lb} \in \mathbb{R}^{12}$ is the conﬁguration of the humanoid manipulator.

We define the configuration $q_{fb}$ of the floating base as a combination of the position of the $\{B\}$ frame origin expressed in $\{I\}$, denoted as $p = \begin{bmatrix}
        p_x & p_y &p_z
    \end{bmatrix}^T$, together with the rotation of $\{B\}$ with respect to $\{I\}$, described by Euler angles $\Phi = \begin{bmatrix}
        \phi & \theta & \psi
\end{bmatrix}^T$. That is
\begin{equation}
    q_{fb} = \begin{bmatrix}
        p \\ \Phi
    \end{bmatrix}\,.
\end{equation}
The conﬁguration of the humanoid manipulator $q_{lb} \in \mathbb{R}^{12}$ is composed of three parts 
\begin{equation}
    q_{lb} = \begin{bmatrix}
        q_{la} \\ q_{ra} \\ q_{h}
    \end{bmatrix}\,,
\end{equation}
 where $q_{la} \in \mathbb{R}^{5}$ and $q_{ra} \in \mathbb{R}^{5}$ represent the joint variables of the 5-DoF left and right arms, $q_{h} \in \mathbb{R}^{2}$ represents the joint variables of the 2-DoF head.


\subsection{Dynamics}


Let $m_i$ be the mass of the $i$-th humanoid torso component, $m_{fb}$ and $m_{rod}$ be the mass of the drone and the suspension base attached to the drone.
The total mass of the system $m$ is
\begin{equation}
    m = m_{fb} + m_{rod} + \sum m_i \;.
\end{equation}
The position of the CoM of the humanoid torso components can be obtained from direct kinematics and is dependent on current humanoid configuration $q_{lb}$.
Defined $c$ as CoM vector of the whole body expressed in the body frame, it holds
 \begin{equation}
c(q_{lb}) =\frac{1}{m}( m_{fb}p_{fb} + m_{rod}p_{rod} + \sum m_ip_{i}(q_{lb}))\; .
\end{equation}
Similarly, the inertial tensor of the platform expressed in the body frame relative to the origin $O'$ can be written as
\begin{equation}
   I_{O'}(q_{lb}) = I_{fb} + I_{rod} +\sum I_{i}(q_{lb})\; .
\end{equation} 

Due to the coupling effect between the aerial platform and the robotic arm, the complete dynamics of the system pose a complex challenge, especially during rapid movements of the robotic arms.
However, serving as an aerial human-operated avatar working in unknown and unstructured working scenarios, rapid arm movements are not part of the design intent of the platform. 
Firstly, rapid arm motion will increase the possibility of accidental collision between the platform and the unknown external environment. 
Secondly, it will also increase the interference of the fuselage movement on the UAV, thereby reducing the overall stability.
Considering the two points above, employing slow arm motion represents as a pragmatic approach to address this problem.
Under the assumption that the robotic arms move at a relatively slow pace, the system dynamics can be simplified as that of a rigid body, allowing for the neglect of coupling effects for practical considerations. This simplification facilitates a more manageable analysis and control of the system.

We can express the dynamics of the body frame as 
\begin{equation}\label{eq:full_dynamics}
\begin{bmatrix}
    F\\ \tau_{O'}
\end{bmatrix} =    
\begin{bmatrix}
    mI_3& -mS(c)\\ mS(c)& I_{O'}
\end{bmatrix} \cdot
\begin{bmatrix}
    \ddot{p}\\ \dot{\omega}
\end{bmatrix}+
\begin{bmatrix}
    -mS^T(\omega)S(\omega)c\\ S(\omega)I_{O'}\omega
\end{bmatrix} \, ,
\end{equation}
where $F$ is the force and $\tau_{O'}$ is the torque applied to the body center $O'$, $S(\cdot)$ is the skew-symmetric matrix operator,
$\omega \in \mathbb{R}^3$ is the angular velocity expressed in the body frame. Let $R \in SO(3)$ be the rotation matrix of the body-ﬁxed frame with respect to the inertial frame, there is
\begin{equation}
\dot{R} =  R S\left(\omega\right)\, .
\end{equation}

Assuming $T_{s}\in \mathbb{R}$ and $M\in \mathbb{R}^3$ to be the total thrust and the total moment generated by the hexacopter motors and taking into account the known compensating force $F_t$ introduced by the tether, $F_{ext}$ is the external force applied to the manipulator and $\tau_{ext}$ is its torque to the body center, the force and torque applied to the body frame are
\begin{subequations} \label{eq:ext_forces} 
\begin{align}
F &= T_{s}Re_3 + F_t - mge_3 + F_{ext} \,,\\
\tau_{O'} &= M  + S(le_3) R^TF_t - S(c) R^T mge_3 + \tau_{ext} \,,
\end{align} 
\end{subequations} 
where  $e_3 = \begin{bmatrix}0 & 0& 1 \end{bmatrix}^T$, and $l$ is the distance from the cable attached point to the origin of the body-fixed frame. 




\section{Control Structure} \label{sec:Control}

The control framework of the aerial platform includes the control system and the teleoperation user interface.

\subsection{Control System}\label{sec:cont_algor}
The control system is divided into three parts, respectively the aerial base, the dual robotic arms, and the head.


\subsubsection{Aerial Base}
The primary control objective for the aerial base is to sustain its position and attitude stability while mitigating interference induced by arm motion and contact when the humanoid torso manipulates. Substituting \eqref{eq:ext_forces} into \eqref{eq:full_dynamics}, six equations arise with six variables, namely $T_{s} \in \mathbb{R}$, $M \in \mathbb{R}^3$, and a two-dimensional direction of $F_{t}$ with a known absolute value. Given the external force $F_{ext}$ and its torque $\tau_{ext}$, a solution to the dynamic equations consistently exists to uphold the equilibrium of the system.

To initiate the analysis, we consider the external force as a disturbance and concentrate on a configuration where the aerial platform stabilizes at a specific point while maintaining a zero attitude.
This configuration is characterized by $p = 0$, $\dot{p} = 0$, $\Phi = 0$, and $\dot{\Phi} = 0$, signifying that the platform's translational velocity, roll angle, and roll rate are all zero. This configuration is considered optimal for various operational scenarios and warrants special attention due to its unique characteristics.

It is known that under small angle assumption, there is $\omega \approx \dot{\Phi} = 0$.
In the neighborhood of this equilibrium, the Coriolis term can be neglected and the dynamic equations in \eqref{eq:full_dynamics} can be written as
\begin{subequations}\label{eq:dynamics}
\begin{align}
T_{s}e_3 + F_t - mge_3  &= m\ddot{p} -mS(c)\dot{\omega} \label{eq:translation_full} \,, \\
 M - S(c) mge_3 &= mS(c)\ddot{p}+ I_{O'}\dot{\omega} \,.
\end{align}
\end{subequations}
In the equilibrium, it holds
\begin{subequations}\label{eq:zero_equilibrium}
\begin{align}
\bar{T}_{s}e_3 + F_t -mge_3 &= 0 \,,\\
\bar{M} - S(c) mge_3 &= 0 \,.
\end{align}
\end{subequations}

Assuming $T_{s_d} = \bar{T_{s}}+ \delta T_{s}$ and $M_d = \bar{M}+ \delta M$ to be the desired total pulling force and total moment generated by the thrusters, substituting it to \eqref{eq:dynamics} and \eqref{eq:zero_equilibrium}, the incremental dynamic equations can be obtained
\begin{subequations}
\begin{align}
\delta T_{s}e_3 &= m\ddot{p} -mS(c)\dot{\omega} \label{eq:translation_dynamic} \,,\\ 
\delta M &= mS(c) \ddot{p}+  I_{O'}\dot{\omega} \label{eq:rotaion_dynamic} \,.
\end{align}
\end{subequations}

In the context of an under-actuated UAV employing parallel thrusters, it is widely recognized that the translation dynamics associated with the $x$ and $y$ directions are coupled with its pitch and roll dynamics. However, the yaw dynamics can be treated as decoupled from the other aspects, and controlling yaw holds a distinct priority \cite{9197055}\cite{lee2010geometric}.
Consider the decomposition of \eqref{eq:translation_dynamic} along $z$ axis, the altitude dynamic can be decoupled as 
\begin{equation}
    \delta T_{s} = m\ddot{p_z} \,.
\end{equation}
It can be derived from \eqref{eq:translation_full} that $p_x$ and $p_y$ will converge to zero when pitch and roll are stabilized due to the suspending force $F_t$. Thus we ignore the translation coupling term in \eqref{eq:rotaion_dynamic} when considering the rotation control.
We expect that the behavior of the system with respect to altitude and attitude manifests in a second-order form, as outlined below
\begin{subequations}
\begin{align}
    \ddot{p_z}+b_1\dot{p_z}+k_1(p_z-p_{z_d})=0 \, ,\\
    \dot{\omega}  + B_2\dot{\Phi} + K_2(\Phi - \Phi_d) = 0 \,.
\end{align}
\end{subequations}
The control law is designed as
\begin{subequations} \label{eq:F_M}
\begin{align}
T_{s_{d}} &=  mg - F_t + m(-b_1\dot{p_z}-k_1(p_z-p_{z_d})) \,,\\
M_{d} &= S(c) mge_3 + I_{O'}(-B_2\dot{\Phi} - K_2(\Phi - \Phi_d)) \,.
\end{align}
\end{subequations}
where $p_{z_d}$ is the reference altitude and $\Phi_d$ is the reference attitude input by a motion planner receiving user commands.

For common commercial UAV platforms with non-open autopilots, such as DJI, access to the control of individual motors is typically restricted. Instead, attitude control is usually achieved through a built-in attitude controller, utilizing the total pulling force of the motor $T_{s_{d}}$ and the desired attitude angle $\Phi_{d}$ or angular velocity $\dot{\Phi}_{d}$.
Consequently, the proposed approach, involving the output of desired $T_{s_{d}}$ and $M_{d}$, cannot be directly implemented but serves as an outer loop connected to the built-in attitude controller. It is assumed that the built-in attitude controller is based on the classic model-independent PD control law, as presented in \cite{90228}. In angular velocity control mode, the relationship between angular velocity command $\dot{\Phi}_{d}$ and the generated torque $\hat{M}$ can be written as:
\begin{equation}
    \hat{M} = b_{\Phi}(\dot{\Phi}_{d}-\dot{\Phi}) \,.
\end{equation}
Substituting $\hat{M}$ with the $M_d$ in \eqref{eq:F_M}, the outer loop control output is expressed as
\begin{subequations}\label{eq:DJI_control}
\begin{align}
T_{s_{d}} &= mg - F_t + m(-b_1\dot{p}_z-k_1(p_z-p_{z_d})) \,,\\
\dot{\Phi}_{d} &= \frac{ S(c) mge_3 + I_{O'}(-B_2\dot{\Phi} - K_2(\Phi - \Phi_d))}{b_{\Phi}} +\dot{\Phi} \,.
\end{align}
\end{subequations}


\subsubsection{Dual Arms}
The control of the arms is divided into two parts. The first computes the position of the joints to obtain the tracking of the reference trajectory. The latter computes the necessary conversion from the joint angles to the motor routing derived from the tendons structure.

The joint position is calculated through a weighted inverse kinematic algorithm \cite{siciliano2010robotics}. 
The desired joint velocities vector of the arm $\dot{q}_{a_d}$ ($\dot{q}_{ra_d}$ for the right arm or $\dot{q}_{la_d}$ for the left) can be computed by
\begin{equation}
    \dot{q}_{a_d} = J_w^\dagger K (x_d-x_e) \,,
\end{equation}
where $K$ is a diagonal coefficient matrix, $x_d$ is the desired end-effector position given by user joysticks, and $x_e$ is the actual end-effector position calculated using direct kinematics function. The matrix $J_w^{\dagger}$ is the weighted pseudo-inverse of the Jacobian matrix $J$:
\begin{equation}
    J_w^{\dagger} = W^{-1}J^T(JW^{-1}J^T)^{-1} \,,
\end{equation}
where $W$ is a symmetric positive definite weight matrix used to optimize the controller performance when the arms are close to a kinematic singularity.

The desired joint variable vector $q_{a_d}$ is calculated using an integration performed in discrete time with numerical techniques:
\begin{equation}
    {q_{a_d}}(t_{k+1}) = {q_{a_d}}(t_k) + \dot{q}_{a_d}(t_k) \Delta t \,.
\end{equation}

The desired motor position command $q_{m_d}$ is obtained from the desired joint variable using the previously specified relationship between joints and tendons:
\begin{equation}
    q_{m_d} = F^{\dagger}q_{a_d} \,,
\end{equation}
where $F^{\dagger}$ is the pseudo-inverse of the configuration matrix $F$.
The command $q_{m_d}$ is used as the reference input of the position servo motor and is executed by the lower-level motor driver.

\subsubsection{Neck and head}
The head controller employs a similar concept of inverse kinematics algorithm as the arm. The expected joint variables are calculated using the desired head position from the Oculus headset orientation.
Since the motors actuate the neck joints directly, the computed joint variables are used as the motor references, and a weighting matrix is unnecessary.

\subsection{Teleoperation} \label{sec:cont_tele}

The teleoperation system is implemented using the GCS introduced in Section \ref{sec:electronics}. This GCS incorporates an immersive VR headset for first-person perspective visualization and head orientation tracking, along with two joysticks for user command input.

The raw images captured by the stereo camera are transmitted to the VR headset, enabling 3D visual reconstruction of the environment as seen by the camera. The movements of the human operator, tracked through the headset and two handheld joysticks, are transformed from the Oculus coordinate system into the body coordinate system of the platform. These transformed movements are then employed as the reference input for the robot head and arm control system.
The thumbsticks of both joysticks are utilized for commanding the movement of the aerial base. Specifically, the left-hand thumbstick is employed to adjust the desired altitude, while the right-hand thumbstick is used for giving yaw rotation commands. The trigger on each joystick is employed to control the closure of the robotic hand.

\begin{figure}[h]
    \centering
    \includegraphics[width = 1.0 \linewidth]{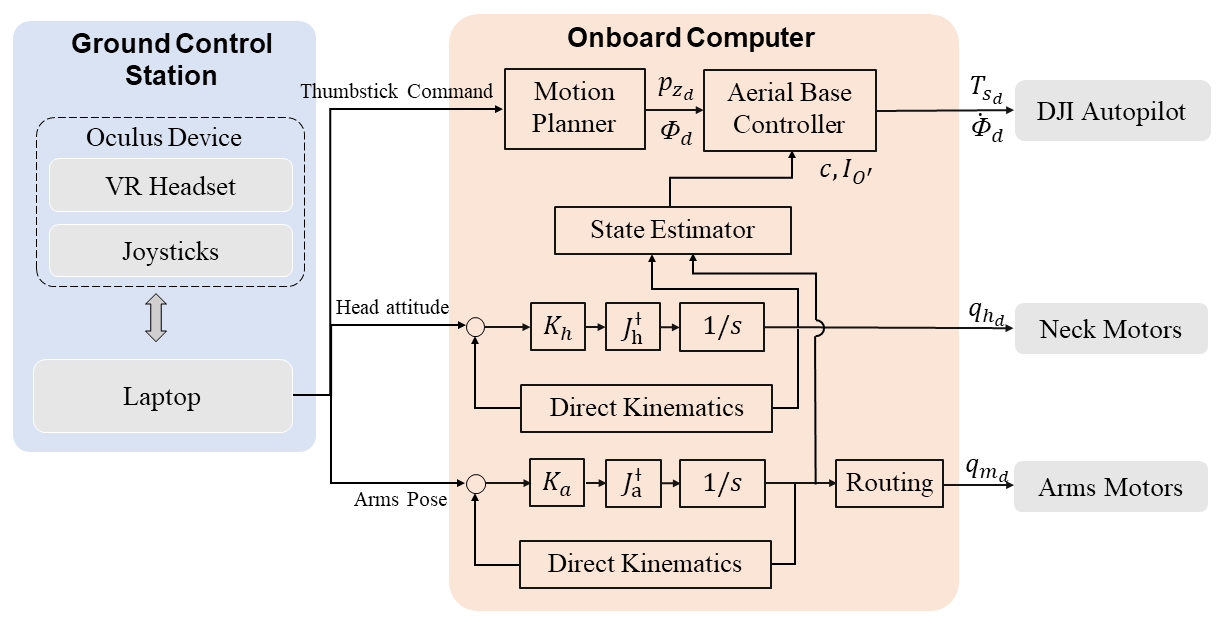}
    \caption{Control Framework. On the left is the teleoperation user interface described in \ref{sec:cont_tele}, in the middle is the control system described in \ref{sec:cont_algor} running on the onboard computer, and on the right is the low-level controller described in \ref{sec:electronics}.}
    \label{fig:control_struct}
    \vspace{0.0cm}
\end{figure}

The GCS software is implemented using the Oculus SDK with a ROS node.
Fig. \ref{fig:control_struct} shows the control framework implemented by ROS running on the onboard computer.
The inputs consist of the thumbstick commands, the arms position and orientation reference trajectory, and finally, the orientation reference for the head. These commands come directly from the teleoperation user interface, described in detail in the previous section (Sec. \ref{sec:cont_tele}).
The output aerial base commands are transmitted to the A3 Pro Flight Controller via a predefined communication protocol in the DJI onboard software development kit. Arm and neck commands are simultaneously sent to their driver board.







\section{Experimental Validation and Results} \label{sec:Results}
The effectiveness of the aerial platform have been validated through two types of experiments.
The performance of the floating base stabilization control described in Sec. \ref{sec:Control} is evaluated by varying the CoM of the system through a series of body movements.
The capabilities of the design are demonstrated through a sequence of tasks to relight light bulbs in a disaster response scenario reproduced under laboratory conditions.

\subsection{Setup} \label{sec:experiment_setup}
\begin{figure}[htbp]
    \centering
    \includegraphics[width = 0.99 \linewidth, trim = 2pt 2pt 2pt 2pt, clip]{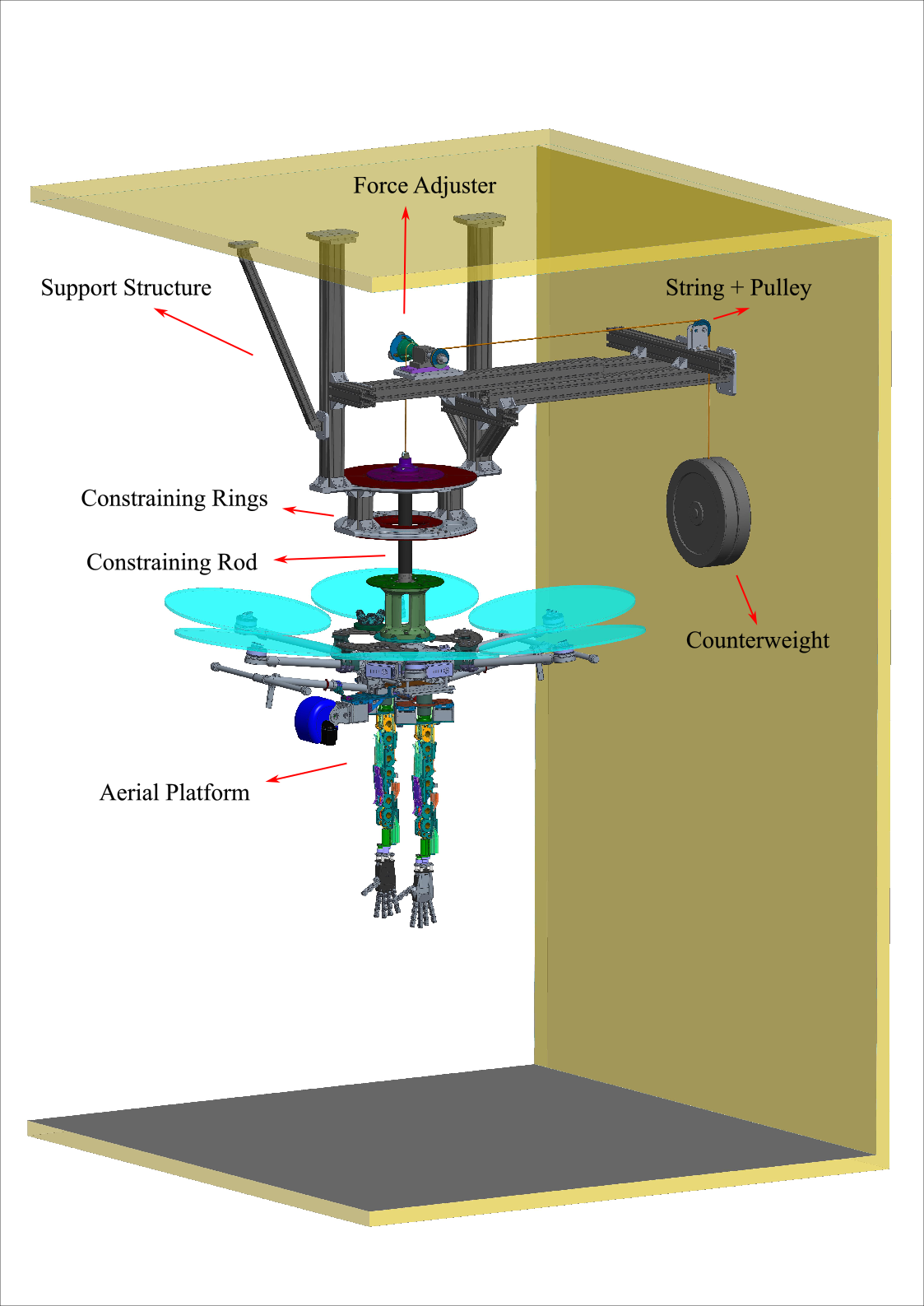}
    \caption{Experiment Setup. The aerial platform is suspended by a wire and pulleys, with a counterweight on the other end that partially compensates for its gravity. A test rig attached to the ceiling and the wall constrains the movement of the aerial platform from accidental collision, described in detail in \ref{sec:experiment_setup}.}
    \label{fig:experiment_setup}
\end{figure}

For safety reasons, all of the experiments were conducted indoors within the protection of a test rig setup. An Optitrack infrared motion capture system was used to provide information for state estimation to restore the outdoor situation where RTK is available.
Fig. \ref{fig:experiment_setup} shows the experiment setup.

The test rig incorporates gravity compensation as described in Sec. \ref{sec:Design}, while limiting the space available for movement of the aerial platform to prevent accidents during flight testing that could cause damage to the human body or the system itself.
The test rig consists of three main parts. The first part is the support structure that holds the entire rig frame to the ceiling.
The second part is a frame constructed from two co-axial parallel rings fixed to the first part, which defines a cylindrical movable space for the platform. It creates setbacks in roll and pitch angles and the radial displacement. 
The third part is a rod-like structure solidly attached to the hexacopter aerial base, adding a weight of $m_{rod} \approx 7$ kg to the main part of the platform of $m_{fb} + \sum m_i \approx 17$ kg. Its upper tip passes through the ring of the second part and restricts the movement of the aerial base in the vertical direction by means of two discs attached to the rod. 
The combination of the constraining components creates an activity space which allows the platform to translate within a cylindrical space with radius $ r \approx 4.7$ cm and height $h \approx 17$ cm, while allowing a rotation of up to $\Delta \phi = \Delta \theta \approx \pm 30^{\circ}$ for pitch and roll, and $\Delta \psi = \pm 360^{\circ}$ for yaw.
More details about the design of the test rig can be found in \cite{IRIM}.

Two gravity compensation schemes based on counterweights and springs were implemented. In the following indoor experiments, the simple counterweight design was deployed. A 15 kg counterweight was used as the force generator, and the force adjuster ratio was set to 1:1.
Considering that the total weight of the aerial manipulator connected to the tether is approximately 24 kg and the tether is configured to compensate for 15 kg, 9 kg of thrust is still to be generated by the aerial base, that is 27\% of its maximum load capacity.
The designed compensating force aims to keep the thrust at a value slightly lower than the nominal state of the hexacopter to reduce the energy consumption of the aerial system.
We chose not to compensate all of the force to prevent the capabilities of the platform to generate rotational torque from declining too much due to insufficient thrust.
In all designed experiments, the aerial platform does not involve rapid motion, and the impact of the additional inertia brought by the counterweight on the system is negligible. Therefore, the compensation force is approximated by the weight of the counterweight $F_t \approx m_cg$, which was validated by the posterior experimental data (in normal working conditions, $\lvert a_c \rvert <0.7 m/s^2 \ll g$ ).

\subsection{Control Validation}

To validate the effectiveness of the proposed control in stabilizing attitude, we conducted two experiments to compare the control performance with and without the proposed aerial base outer loop control in a disturbance rejection task.

As mentioned above, the DJI autopilot requires additional sensor support for hovering in an indoor scenario. Therefore, the altitude and yaw control of the aerial base described by \eqref{eq:DJI_control} was facilitated to keep the aerial manipulator hovering in both of the two experiments, wherein Optitrack feedback was employed.
The joints of the arms are systematically controlled following a predetermined motion sequence to induce disturbances in the system.
The fundamental distinction between the two experiments lay in the control of roll and pitch. In one experiment, solely the DJI built-in attitude controller was utilized with zero as the input for roll and pitch control. In the other experiment, the proposed outer loop control was applied for roll and pitch.
The uniform objective remained consistent with stabilizing body attitude angles $\Phi$ at the zero position. This goal served as the benchmark for evaluating the performance of the proposed controller in mitigating disturbances induced by CoM shifting and effectively stabilizing the desired attitude.

\begin{figure*}[ht]
\centering
    \centering
        \includegraphics[width = \linewidth, trim=0cm 0 0cm 0, clip]{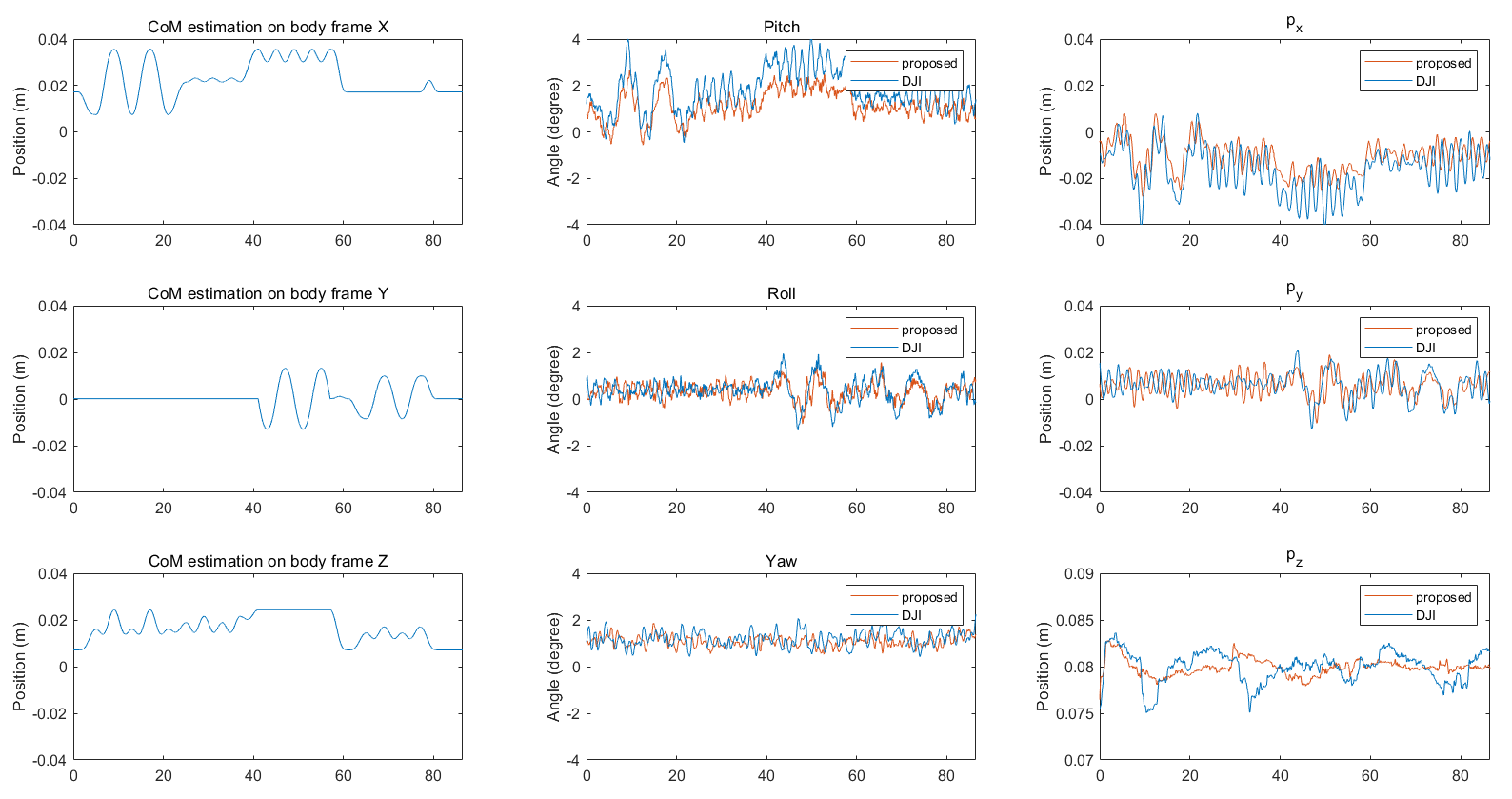}
    \caption{Plots of control validation experiment result. 
    Response of the two controllers to a test shift in the Center of Mass position (left column). Pitch, Roll, and Yaw (central column) and $p_x$, $p_y$ and $p_z$ positions of the body frame (right column). The DJI controller is in blue, and the proposed controller is in red.
    }
    \label{fig:control_validation}
    \vspace{0.0cm}
\end{figure*}

The results are shown in Fig.~\ref{fig:control_validation}. 
An estimate of the overall CoM shift caused by the movement of the robotic arm is illustrated in the figure.
It can be seen (left column of Fig.~\ref{fig:control_validation}) that the larger CoM variation happens along the x-direction of the body.
From the center and right column of  Fig.~\ref{fig:control_validation}, it is possible to appreciate that applying the proposed controller can compensate the disturbance induced on the body attitude better than with only the DJI build-in controller.
The root mean square (RMS) error of the airframe attitude and position under the two control are compared in Table ~\ref{tab:control_validation}.
Along the X-direction, where the disturbance is the most significant, the RMS error of the airframe attitude under the control of the proposed controller is 1.3218, compared to 2.0367 for the DJI controller.

It is observed that the oscillations in position are relatively small, on the scale of centimeters, and the differences between the experiments are not significant. This is attributed to the relatively high system stability of the tethered system.
The dynamic model indicates that the position of the platform is coupled with its attitude due to the under-actuated hexacopter aerial base. When the suspension point is stationary, the position deviation tends to converge to zero as the attitude converges to zero. However, if the suspension point moves, the dynamics will differ, requiring further analysis. Nonetheless, this situation involves the motion of the airborne carrier and is beyond the scope of this paper.

\begin{table}[htbp]
    \caption{Performance comparison between proposed control versus DJI control.}
    \label{tab:control_validation}
    \centering
\begin{tabular}{|p{35pt}|p{20pt}|p{20pt}|p{20pt}|p{20pt}|p{20pt}|p{20pt}|}
\hline
\rowcolor{LightSkyBlue} 
\cellcolor{LightSkyBlue} & \multicolumn{3}{c|}{\cellcolor{LightSkyBlue}Rotation$(^{\circ})$} & \multicolumn{3}{c|}{\cellcolor{LightSkyBlue}Position$(m)$} \\ \cline{2-7} 
\rowcolor{LightSkyBlue} 
\multirow{-2}{*}{\cellcolor{LightSkyBlue}RMS errors}& $\theta$ & $\phi$ & $\psi$ & $p_x$ & $p_y$ & $p_z$  \\ \hline
DJI& $2.0367$  & $0.6338$  & $1.2343$ & $0.0190$ & $0.0082$ &$0.0800$ \\ \hline
\rowcolor{LightGray} 
proposed & $1.3218$ & $0.4986$  & $1.1114$ & $0.0130$ & $0.0074$ &$0.0799$ \\ 
\hline
\end{tabular}
\end{table}

\subsection{Tasks Validation} \label{sec:taskvalidation}
\begin{figure*}[ht]
    \centering
    \includegraphics[width = \linewidth ]{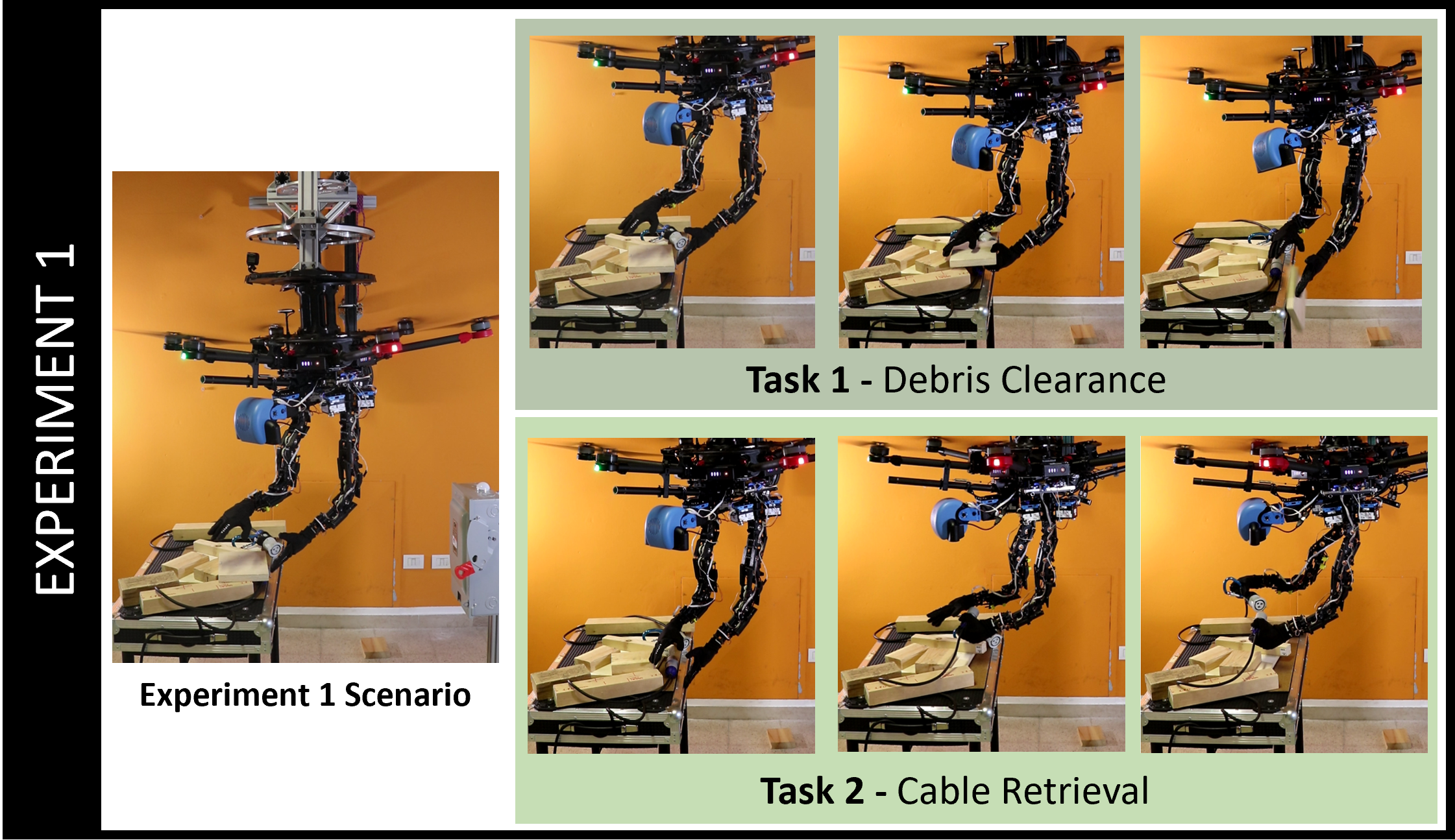}
    \caption{Photo sequence of the first experiment scenario as described in \ref{sec:taskvalidation}, in which the aerial platform sequentially performed two tasks: Debris Clearance and Cable Retrieval. The platform first picked up and removed a wooden brick to reveal the buried aviation connector. Then, it used two hands to pick up the two connectors, one in each hand.}
    \label{fig:Experiment_1}
\end{figure*}

\begin{figure*}[ht]
    \centering
    \includegraphics[width = \linewidth ]{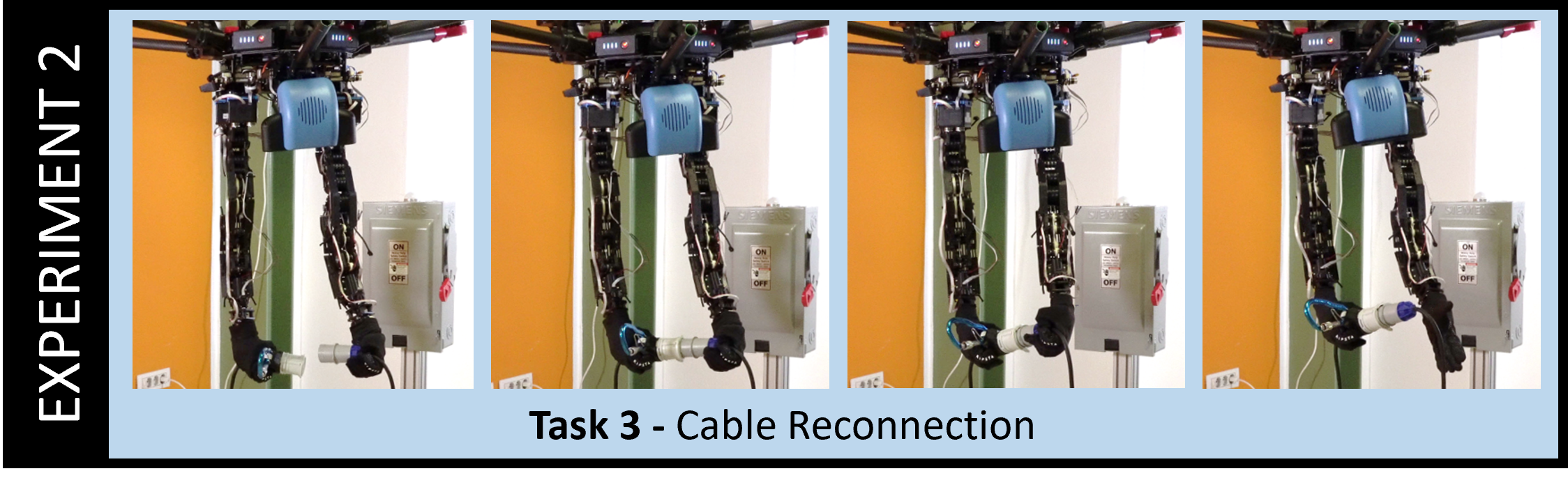}
    \caption{Photo sequence of the second experiment of Cable Reconnection as described in \ref{sec:taskvalidation}. The aerial platform used both hands to collaboratively insert the connector it grabbed in the previous scenario.}
    \label{fig:Experiment_2}
\end{figure*}

\begin{figure*}[ht]
    \centering
    \includegraphics[width = \linewidth ]{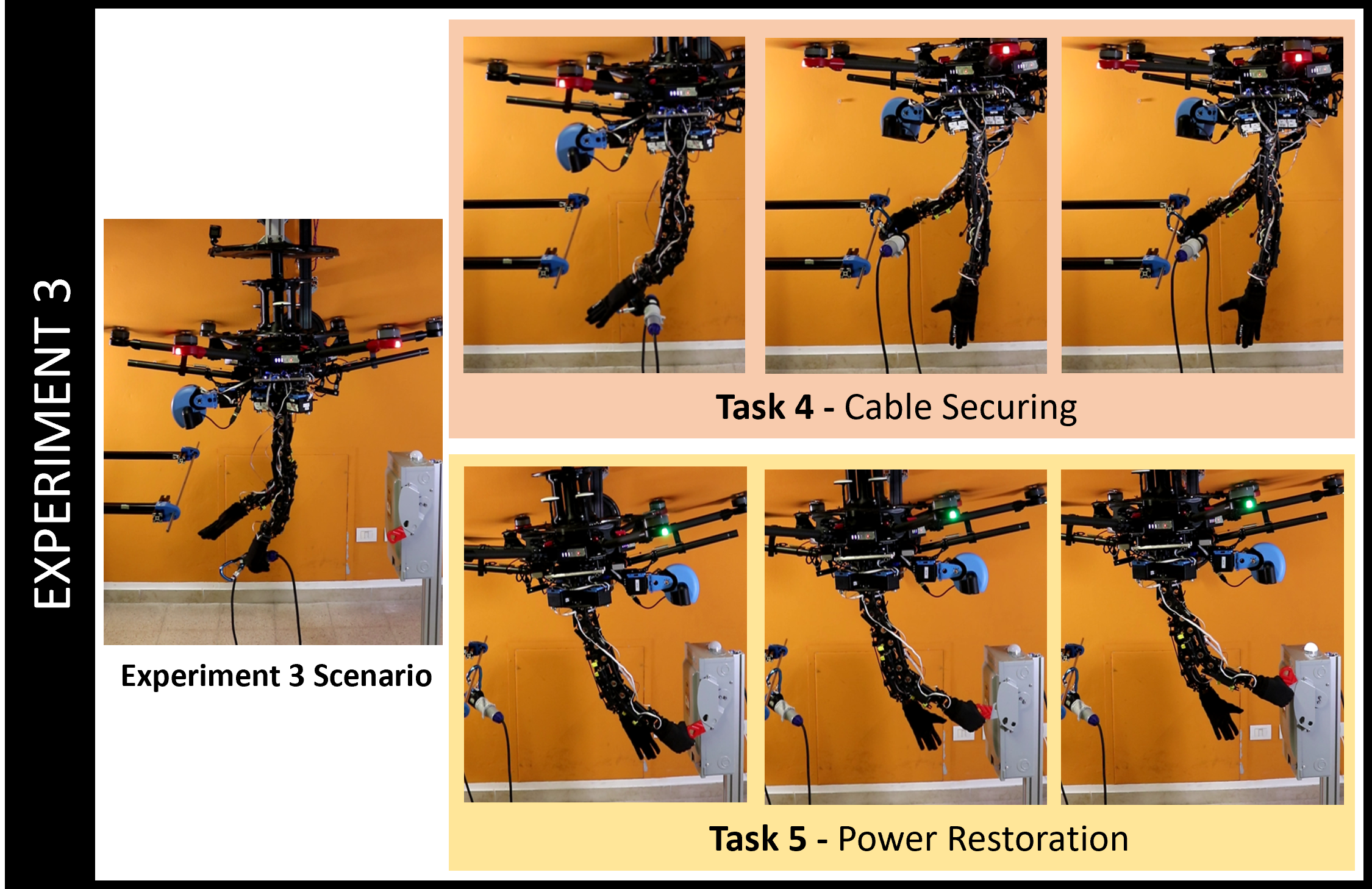}
    \caption{Photo sequence of the third experiment scenario as described in \ref{sec:taskvalidation}, in which the aerial platform sequentially performed Cable Securing and Power Restoration. It first switched the connected connector from the left hand to the right hand, raised the height, and snapped the spring hook attached to the connector onto the crossbar. Then, it turned backward and lifted the switch up to light the bulb.}
    \label{fig:Experiment_3}
\end{figure*}
The capacity of aerial manipulators for flight confers the distinct advantage of disregarding ground-level debris and obstacles, rendering them particularly well-suited for applications in post-disaster scenarios.

To substantiate the overall functionality of the proposed aerial platform, we designed a scenario simulating a post-disaster situation. In this scenario, a severely damaged building has experienced a disruption in its internal power and communication cables. To facilitate rescue and reconstruction, robots are required to access the building, re-establish the cable connections, and restore power.
To facilitate systematic reproduction and analysis within a laboratory setting, we divided the workflow into five tasks:
\begin{itemize}
    \item Debris Clearance: The initial phase involves the clearance of debris at the site to locate the disconnected cables efficiently.
    \item Cable Retrieval: Subsequently, the disconnected cables are gathered for further handling.
    \item Cable Reconnection: Following collection, the cables are skillfully reconnected, necessitating the use of both hands.
    \item Cable Securing: To mitigate the risk of short circuits due to potential ground-level water exposure, the cables are elevated to a higher level by a hook.
    \item Power Restoration: The process culminates in the restoration of power through the activation of a switch.
\end{itemize}

Due to the limitations of the movement space and safety considerations, these five steps could not be accomplished in a single session, so we performed the five tasks in three scenarios. 
Wood bricks of different sizes and shapes on a table are used to represent debris on the ground. a cable connected with industrial aviation connectors are used in cable reconnection experiments. a typical spring hook and electric box are used in the last two experiments.
All the experiments were performed with the control methods described in Sec. \ref{sec:Control} and were teleoperated remotely by a human operator located in another room.

\paragraph{Debris Clearance (1) and Cable Retrieval (2)}

Fig. \ref{fig:Experiment_1} shows the first two tasks completed consecutively in one experiment.
In the experimental scenario, wooden bricks of diverse sizes and shapes were arranged in a random manner on a table to replicate the appearance of scattered debris. Concealed beneath these simulated debris, there was a cable equipped with industrial aviation connectors.
At the beginning of the task, the platform looked around to locate the connectors. Subsequently, the manipulators engaged in the retrieval and removal of the debris covering the connectors. Following this, it proceeded to employ each of its hands to grasp the exposed connectors.

This experiment demonstrates the grasping ability of the proposed aerial platform for objects of different sizes and shapes. The presence of soft hands enables the platform to manipulate irregularly shaped objects.
It is worth mentioning that during the grasping process, the arm unavoidably came into contact with the tabletop and exerted a force. The flexible joints successfully mitigated the impact forces during the contact, and the controller resisted this disturbance and smoothed the body attitude.

\paragraph{Cable Reconnection (3)}
In order to provide a clearer visual representation of this procedure, we removed the table used in the preceding experiment to create the necessary space. Fig. \ref{fig:Experiment_2} presents the process of connecting the grasped aviation connectors. 
Alignment and splicing is a challenging task that requires two-handed collaboration to accomplish.
The arm dexterity and bi-manual manipulation ability are demonstrated in this task.

\paragraph{Cable Securing (4) and Power Restoration (5)}
Fig. \ref{fig:Experiment_3} introduced the last two tasks performed consecutively.
In this scenario, a standard spring hook, which was attached to the connector from the beginning, was used for the hooking task, and a common electric box was employed for the switch task.
The aerial platform is initiated by grasping the connector with its left hand, subsequently transitioning to the spring hook onto its right hand. Following this, it elevated the assembly to the desired height and engaged the spring hook onto a beam structure. Subsequently, it executed a 180-degree rotation and proceeded to vertically actuate the switch, ultimately lighting the bulb.

In a more coordinated motion to lift the switch, the operator elevated the height of the aerial base while simultaneously lifting the arm, effectively leveraging the freedom of movement of the aerial base as retained in our design. The compliance of the arms played a crucial role in successfully executing this task during the procedure.
It is noteworthy that during the initial attempt to lift, the hand grasp was not positioned correctly, resulting in the finger getting caught in the switch. To free the hand from the switch, the arm had to apply a substantial force, consequently inducing a significant disturbance on the body. However, the flexible joint design preserved the integrity of the arms, and the proposed controller ultimately stabilized the body.

\subsection{Discussion}
The experimental results demonstrate the effectiveness of the system in environments that are challenging to access from the ground. It exhibits the necessary dexterity to handle tasks involving different targets, adaptively manipulating objects of varying shapes and sizes. The proposed control and teleoperation framework effectively maintains body stability in flight, demonstrating robustness to disturbances. Simultaneously, it allows the operator to safely and intuitively engage in various tasks involving physical interaction with the external environment.

Furthermore, experiments (4) and (5) showcased the system manipulation capabilities in a cable suspension working environment, illustrating ascent, descent, and rotation capabilities. The manipulation workspace was successfully demonstrated within the constraints of the test rig, suggesting the potential for further expansion beyond these limitations, particularly with the utilization of a movable suspension base, such as a crane.

\subsubsection*{Limitations and future work}
The choice of the DJI M600 Pro as the aerial base is driven by its sufficient drive capabilities. However, it is essential to note that its powerful downwash airflow can pose challenges when dealing with lighter objects such as paper, making their operation difficult.
In light of this, there is a willingness to explore different possibilities by considering the use of alternative multirotors, potentially smaller ones. An investigation into the advantages and limitations of these alternatives will be a focus of further research.
While the tendon transmission design of the arm offers improved weight distribution and compliance, it does come with a trade-off in terms of lower end-effector positioning accuracy compared to direct transmission. However, this deviation can be easily compensated for by human operators during teleoperation, highlighting the adaptability of first-person perspective operation.

The proposed controller exhibits the capability to mitigate the effects of CoM shifts induced by arm movements and displays a degree of resilience against external forces imposed by the environment.
However, it has some limitations: the controller is based on a simplified model, where the reaction torque caused by whole-body motion is not fully considered; during the linearization process, the designed controller primarily focuses on balancing the system at the zero attitude which treats the external force as a disturbance and doesn't take its control into account.
These limitations bring some problems: when the aerial platform performs fast limb movements or encounters large external disturbances, the performance of the proposed controller becomes compromised. 
For example, experimental observations show that under a huge external disturbance, the system may not revert to the initial attitude but instead stabilize at a non-zero attitude.
This phenomenon may be attributed to inherent nonlinearities within the system or other unmodeled phenomena, possibly arising inside the closed DJI hexacopter controller or the unmodeled external force, necessitating further in-depth analysis.
Therefore, our further research aims to enhance its capability and robustness under such conditions, for which a more complex whole-body nonlinear controller is required. 
In the future, the applied force will be further included in the control, for which force estimation may also need to be considered.

One of the main factors that affect the operational effectiveness lies in the absence of haptic sensors and force sensors. Relying solely on visual information hinders the intuitive perception of the state of the operating object. This limitation sometimes leads to misjudgments in the operational process, resulting in operational failures.
Thus, we plan to integrate haptic sensors and force estimation into teleoperation in the future. 
In addition, the video streaming delay caused by the limited transmission speed is also a factor that affects the overall operating experience.
This can be improved with the advancement in communication technology.

\section{Conclusions} \label{sec:Conclusion}


In this paper, we presented a novel Suspended Aerial Manipulation Avatar prototype. It comprises a hexacopter aerial base and a humanoid torso together with a variable-length suspension system, and is capable of safely executing various interactive tasks in unknown environments. 
In comparison to existing solutions, our approach offers an advantage by preserving the freedom of motion for the aerial base while utilizing variable-length cable suspension to extend flight duration. 
Furthermore, concerning the design of the aerial floating base, our solution is universal as it can be integrated with commercial UAVs instead of necessitating a dedicated design.
A stabilizing control that considers the humanoid torso motion and the role of the tethering system, complemented by an immersive teleoperation framework was presented.
The SAM-A prototype and the proposed control were demonstrated in experiments, highlighting its strengths and weaknesses.

Our further research will focus on a more comprehensive nonlinear whole-body controller to explore the potential of the platform.
Future investigations will also explore the application of the aerial platform in field-like experiments, where the aerial manipulation avatar could operate attached to a large crane and/or in free-flight mode.
Moreover, we will investigate the deployment of autonomous operation modes on the aerial platform to assist and simplify teleoperation, as well as further achieve full autonomous control.





\section*{Acknowledgment}
The authors would like to thank Giacomo Dinuzzi, Manuel Barbarossa for their help with the arm design, Andrea Di Basco, Vinicio Tincani, Mattia Poggiani and  Cristiano Petrocelli for their valuable support in the development of the hardware prototypes, and Andrea Cavaliere, Federica Barontini for their assistance in the experimental validation.


\bibliographystyle{IEEEtran}
\bibliography{IEEEexample}


\end{document}